\documentclass{article}

\usepackage{microtype}
\usepackage{subfigure}
\usepackage{booktabs} 
\usepackage{wrapfig}

\usepackage{hyperref}
\usepackage{array} 
\usepackage{booktabs}
\usepackage{multirow}
\usepackage[outline]{contour}
\usepackage[normalem]{ulem}
\useunder{\uline}{\ul}{}
\usepackage[utf8]{inputenc} 
\usepackage[T1]{fontenc}    
\usepackage{hyperref}       
\usepackage{url}            
\usepackage{booktabs}       
\usepackage{amsfonts}       
\usepackage{nicefrac}       
\usepackage{microtype}      
\usepackage[dvipsnames]{xcolor}
\usepackage{multirow}
\usepackage{array}
\usepackage{graphicx}
\usepackage{subfigure}
\usepackage{float}

\usepackage[accepted]{icml2024_grg}

\usepackage{amsmath}
\usepackage{amssymb}
\usepackage{mathtools}
\usepackage{amsthm}

\usepackage{algpseudocode}

\usepackage[capitalize,noabbrev]{cleveref}

\theoremstyle{plain}

\theoremstyle{definition}

\theoremstyle{remark}

\usepackage[textsize=tiny]{todonotes}

\icmltitlerunning{LCA-on-the-Line: Benchmarking Out of Distribution Generalization with Class Taxonomies}

\begin{document}

\twocolumn[
\icmltitle{\emph{LCA-on-the-Line}: Benchmarking Out-of-Distribution \\ Generalization with Class Taxonomies}



\icmlsetsymbol{equal}{*}
\begin{icmlauthorlist}
\icmlauthor{Jia Shi}{cmu}
\icmlauthor{Gautam Gare}{cmu}
\icmlauthor{Jinjin Tian}{cmu}
\icmlauthor{Siqi Chai}{cmu}
\icmlauthor{Zhiqiu Lin}{cmu}
\icmlauthor{Arun Vasudevan}{cmu}
\icmlauthor{Di Feng}{argo,apple}
\icmlauthor{Francesco Ferroni}{argo,nvidia}
\icmlauthor{Shu Kong}{tam,mau}
\end{icmlauthorlist}

\icmlaffiliation{cmu}{Carnegie Mellon University}
\icmlaffiliation{tam}{Texas A\&M University}
\icmlaffiliation{mau}{University of Macau}
\icmlaffiliation{nvidia}{Now at Nvidia}
\icmlaffiliation{apple}{Now at Apple}

\icmlaffiliation{argo}{Work done at Argo AI GmbH}
\icmlcorrespondingauthor{Jia Shi}{jiashi@alumni.cmu.edu}
\icmlcorrespondingauthor{Shu Kong}{skong@um.edu.mo}


\icmlkeywords{Deep Learning, Computer Vision, Deep Neural Networks, Image Classification, Image Recognition, ImageNet Dataset, Causal Learning, Hierarchical Classification, Cross-Domain Generalization, Transfer Learning, Spurious Correlation, Robustness Evaluation, Model Generalization, Out-of-Distribution Generalization, Representation Learning, Hierarchical Distance, Taxonomy, Vision-Language Models, Class Taxonomy, Zero-Shot Learning, Benchmarking, Invariant Risk Minimization, Feature Transferability, ICML}

\vskip 0.3in
]



\printAffiliationsAndNotice{}  

\begin{abstract}
We tackle the challenge of predicting models' Out-of-Distribution (OOD) performance using in-distribution (ID) measurements without requiring OOD data. Existing evaluations with ``Effective Robustness'', which use ID accuracy as an indicator of OOD accuracy, encounter limitations when models are trained with diverse supervision and distributions, such as class labels (\textit{Vision Models, VMs, on ImageNet}) and textual descriptions (\textit{Visual-Language Models, VLMs, on LAION}). VLMs often generalize better to OOD data than VMs despite having similar or lower ID performance. To improve the prediction of models' OOD performance from ID measurements, we introduce the \emph{Lowest Common Ancestor (LCA)-on-the-Line} framework. This approach revisits the established concept of LCA distance, which measures the hierarchical distance between labels and predictions within a predefined class hierarchy, such as WordNet. We assess 75 models using ImageNet as the ID dataset and five significantly shifted OOD variants, uncovering a strong linear correlation between ID LCA distance and OOD top-1 accuracy. Our method provides a compelling alternative for understanding why VLMs tend to generalize better. Additionally, we propose a technique to construct a taxonomic hierarchy on any dataset using $K$-means clustering, demonstrating that LCA distance is robust to the constructed taxonomic hierarchy. Moreover, we demonstrate that aligning model predictions with class taxonomies, through soft labels or prompt engineering, can enhance model generalization. Open source code in our \href{https://elvishelvis.github.io/papers/lca/}{Project Page}.

\end{abstract}

\begin{figure}[h]
\centering
\includegraphics[width=\columnwidth]{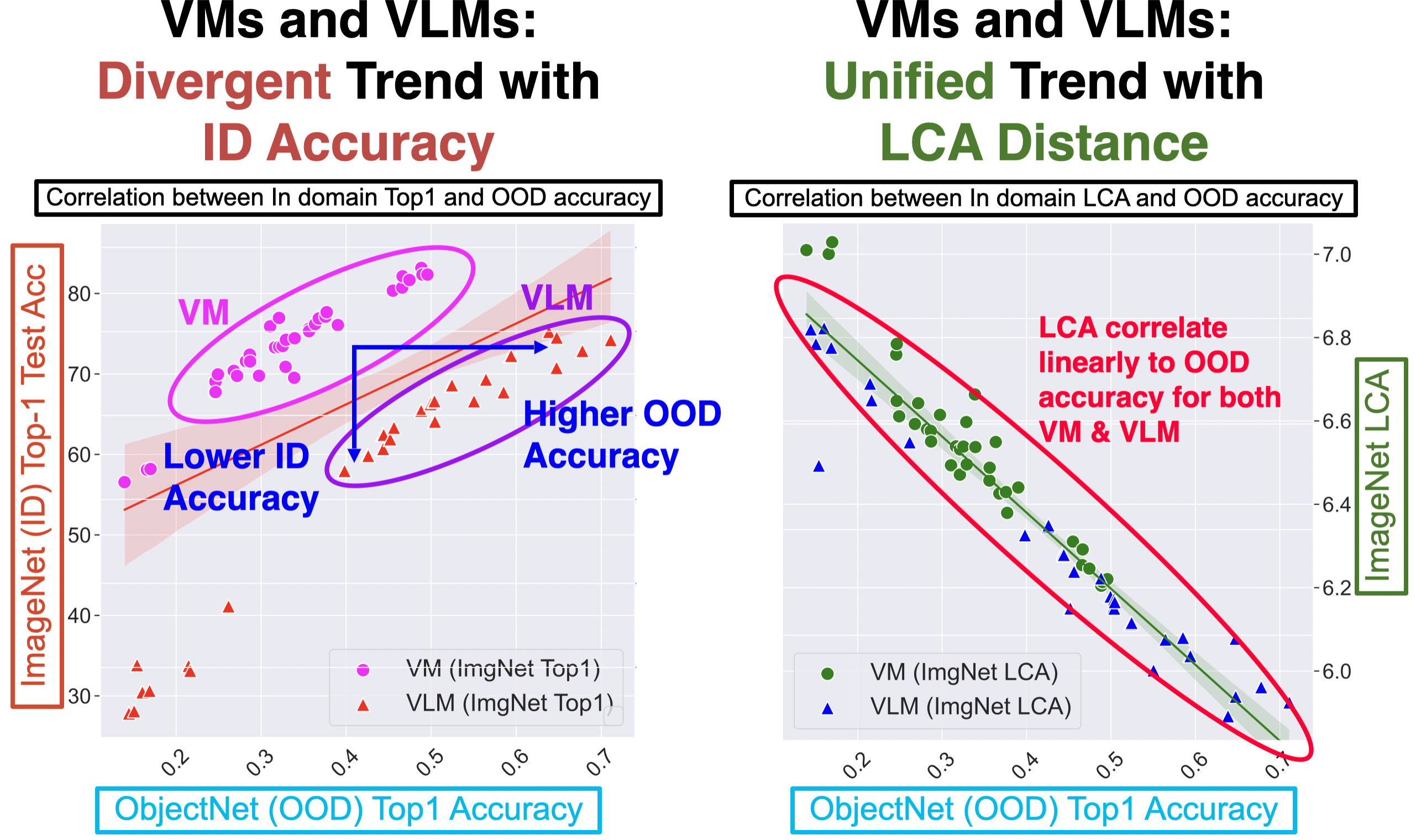}
\vspace{-4mm}
\caption{\textbf{Correlation between LCA distance and out-of-distribution (OOD) performance in Vision and Vision-Language Models (VLMs).} 
In both panels, the \textcolor{cyan}{X-axis} represents the top-1 accuracy on ObjectNet (OOD test dataset). The Y-axes depict the \textcolor{Red}{top-1 accuracy (left-axis)} and 
\textcolor{LimeGreen}{LCA distance (right-axis)} on ImageNet (ID test dataset). 
The left plot reveals a divergent trend where Vision Models (VMs) show a trade-off between OOD and \textcolor{Red}{ID accuracy}, while VLMs tend to maintain higher OOD accuracy regardless of ID performance. 
The right plot demonstrates a unified, strong positive correlation between \textcolor{LimeGreen}{LCA distance} and OOD accuracy for both VMs and VLMs, showing that LCA distance is a robust metric for evaluating model generalization across different architectures, model modalities, and training data sources.
}
\vspace{-2mm}
\label{fig:explain_vlm}
\end{figure}

\begin{figure*}[t]
\centering
 \includegraphics[width=0.98\textwidth]{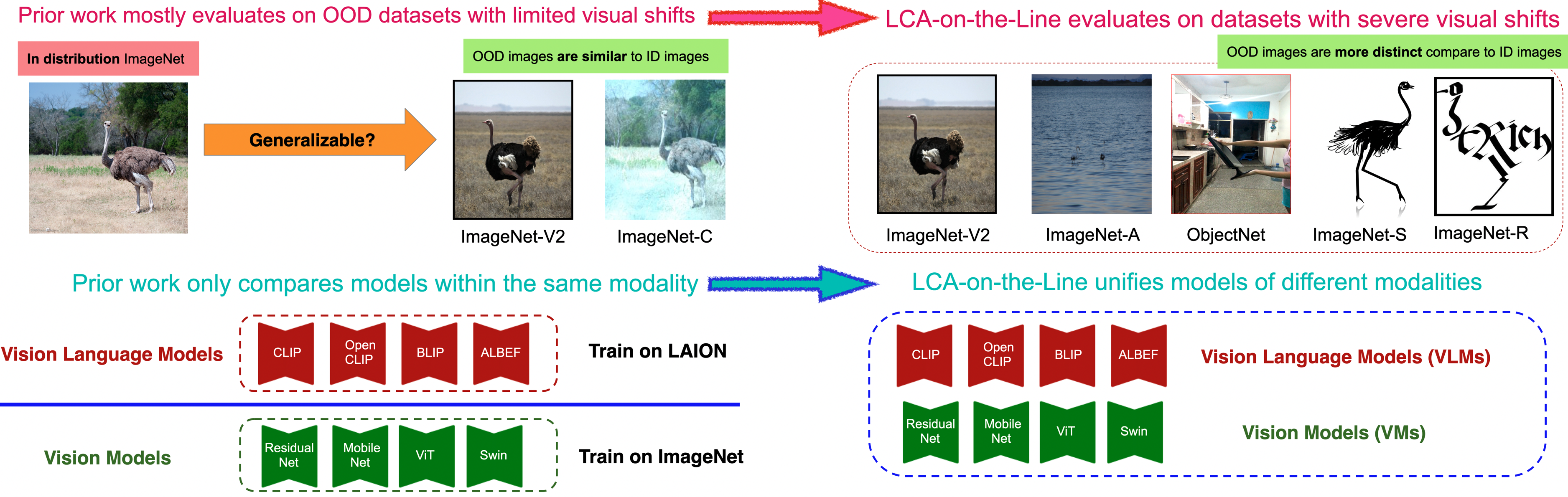}
 \vspace{-3mm}
   \caption{\textbf{Comparison of our setting with prior work.} \textbf{Left}: prior work settings such as Accuracy-on-the-line~\cite{miller2021accuracy} and Agreement-on-the-line~\cite{baek2022agreement}.
   \textbf{Right}: our setting. To the best of our knowledge, LCA-on-the-line is the first approach to uniformly measure model robustness across VMs and VLMs on OOD datasets with significant distribution shifts (ImageNet-S/R/A/O).}
\label{fig:setting_compare}
\end{figure*}

\section{Introduction}
Generalizing models trained on in-distribution (ID) data to out-of-distribution (OOD) conditions is a notoriously difficult task. Distribution shifts undermine the independent and identically distributed (IID) assumption between training and testing data, challenging the model's robustness. Numerous OOD datasets have been proposed to study the effects of different interventions, such as temporal shifts~\cite{hu2022drinking, lomonaco2017core50, lin2021clear}, artificial noise~\cite{hendrycks2019benchmarking, arjovsky2019invariant, larochelle2008zero}, and natural distribution shifts~\cite{hendrycks2021many, hendrycks2019benchmarking,barbu2019objectnet,recht2019imagenet}. Maintaining model robustness becomes significantly more difficult with severe visual shifts in the image domain. However, many studies evaluate generalization on OOD datasets with limited visual shifts or only involve artificial noise, such as ImageNet-v2 or ImageNet-C \cite{recht2019imagenet, arjovsky2019invariant}. Such datasets fail to fully reflect a model's generalization capability when confronted with severe distribution shifts \cite{hendrycks2021many, hendrycks2019benchmarking, barbu2019objectnet}, as there is often limited transfer of robustness from synthetic to natural distribution shifts \cite{taori2020measuring}.


In the realm of model generalization, numerous attempts have been made to predict a model's performance on OOD datasets based on in-distribution measurements, following the concept of \textit{effective robustness} \cite{taori2020measuring}. These approaches, referred to as `X-on-the-line' \cite{miller2021accuracy, baek2022agreement}, suggest that a model's OOD performance is correlated to in-distribution accuracy~\cite{miller2021accuracy, recht2019imagenet, miller2020effect,roelofs2019meta} or models consensus on in-distribution accuracy~\cite{jiang2021assessing, baek2022agreement}.

Moreover, several prior attempts rely on domain generalization strategies that necessitate prior knowledge of the target domain or require an estimation of OOD domain information~\cite{chen2021mandoline, li2022multi}. These can lead to computationally intensive processes, particularly when involving multiple models or inferences~\cite{baek2022agreement,deng2022strong}.

Most prior research has focused solely on estimating generalization among vision models (VMs) supervised on class labels trained on ImageNet~\cite{taori2020measuring, mustafa2020deep}. Emerging large-scale Vision-Language Models (VLMs) trained on datasets like LAION demonstrate exceptional generalization performance on out-of-distribution (OOD) data. However, as shown on the left plot of Fig.~\ref{fig:explain_vlm}, existing evaluation~\cite{miller2021accuracy} using ID accuracy fail to explain the effective robustness~\cite{taori2020measuring} gap between VMs and VLMs. This underscores the necessity to evaluate and compare models across different families under a unified evaluation framework. Recently, \cite{shi2023effective} observed the same problem and proposed evaluating OOD accuracy using multiple ID test sets, but their method requires multiple evaluation runs. 

Unlike VMs, VLMs leverage more diverse training data, contrastive loss, and language supervision. There have been attempts to measure VLM generalization~\cite{haochen2021provable,fang2022data, schuhmann2022laion, kaur2022modeling}, specifically suggesting that diversity in training data is an indicator of model generalization. 
However, it is non-trivial to measure data diversity, 
and even collect and train on such large-scale diverse data~\cite{schuhmann2022laion}.

Prior attempts lack a unified, simple measurement for both VMs and VLMs to explain model generalization and convert it into actionable improvements. To address the issues of (1) lack of unified metrics for VLMs and VMs, or models trained on different data sources; (2) need for robustness to large domain shifts; (3) desire for computationally efficient metrics, we propose adopting the Lowest Common Ancestor (LCA) distance to measure model generalization. The LCA distance is the taxonomic distance between labels and predictions, given a predefined class hierarchy, such as WordNet. 
Through a series of empirical experiments involving 75 models (36 VMs and 39 VLMs) (cf.  Fig.~\ref{fig:setting_compare}), we show 
that the in-distribution LCA distance \textbf{strongly correlates} with multiple ImageNet-OOD datasets under severe visual shifts (cf. Fig.~\ref{fig:explain_vlm} right plot).
\emph{This finding may help explain the surprising result that zero-shot vision-language models with poor top-1 accuracy generalize better to novel datasets compared to state-of-the-art vision models. This spurs us to further investigate and discuss the potential of the LCA benchmark for improving model generalization.} We also discuss the suitability of LCA as a generalization indicator in Section~\ref{section_hypothesis}.

In summary, we make the following major contributions: (1) We propose the Lowest Common Ancestor (LCA) distance as a new metric for evaluating model generalization. This benchmark utilizes class hierarchies, such as WordNet, which encode relationships between classes. (2) We validate our benchmarking strategy through large-scale experiments, analyzing 75 models across five ImageNet-OOD datasets. Our findings reveal a strong linear correlation between in-distribution LCA and OOD Top-1 performance, thus establishing the `LCA-on-the-Line' framework. (3) We offer a thorough analysis of the connection between LCA and model generalization, providing new insights to inspire further research in this area. (4) For datasets without a predefined hierarchy, we introduce a method for constructing latent hierarchies using K-means clustering. Our results demonstrate that the LCA distance is robust to variations in underlying taxonomies or hierarchies. (5) We illustrate the potential of this benchmark by demonstrating how model generalization can be enhanced by aligning model predictions with the class hierarchy.

\section{LCA Distance Measures Misprediction Severity}

\begin{figure}[t]
\includegraphics[width=\columnwidth]{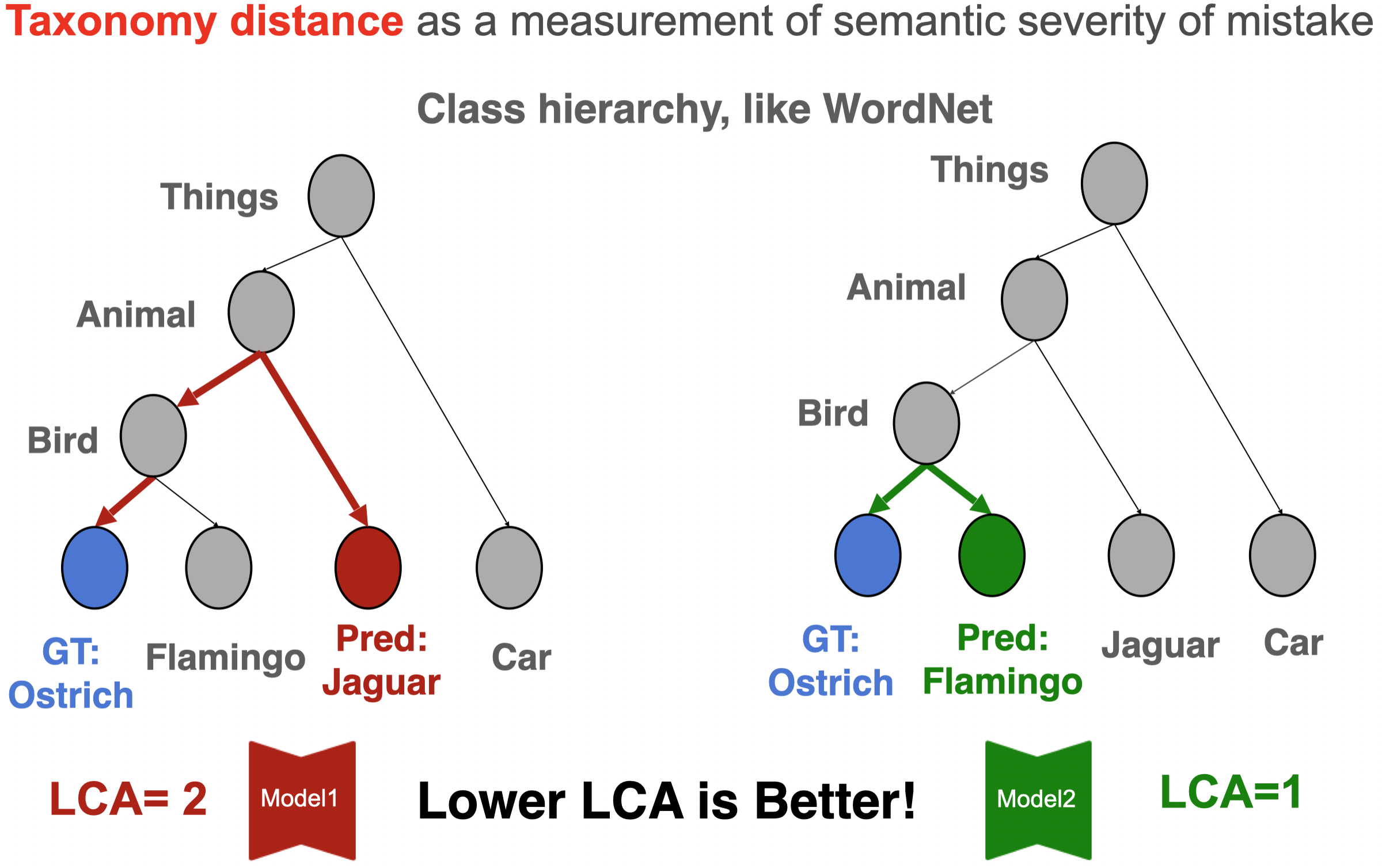}
\vspace{-4mm}
\caption{\textbf{LCA distance visualization}. Our method estimates a model's generalization based on its in-distribution semantic severity of mistakes. We use the `Lowest Common Ancestor' (LCA) distance to rank the distance between the model's prediction and the ground-truth class within a predefined taxonomic hierarchy, such as WordNet. The LCA distance is proportional to the shortest path from the prediction to the ground-truth class in the hierarchy.}
\label{fig:hier}
\end{figure}

We propose using the in-distribution Lowest Common Ancestor (LCA) distance, also known as taxonomy loss, as a predictor for model generalization. Here, we formally define how taxonomy loss can be measured using in-distribution data.
Taxonomy loss measures the class ranking difference between a model's prediction based on class likelihood, and a predefined class order encoded by class taxonomy. Lower taxonomy loss is expected when a model assigns higher likelihood to classes that are semantically closer to the ground-truth class, in other words, `making better mistakes' \cite{bertinetto2020making, peri2023towards}. For example, if a cat image is predicted as a dog by model-A and as a car by model-B, model-A would have a lower LCA distance as it makes a better mistake than model-B. Following previous research \cite{bertinetto2020making, imagenet_cvpr09}, we use WordNet \cite{miller1990introduction}, a large-scale lexical database inspired by psycholinguistic theories of human lexical memory \cite{miller1995wordnet}, to encode class taxonomy. The WordNet taxonomy is well suited for the widely used ImageNet dataset which builds on WordNet. An example of LCA distance is shown in Fig~\ref{fig:hier}.

Given two classes, $y$ (the ground-truth class) and $y'$ (the prediction class), we define the \textbf{LCA distance} according to \cite{bertinetto2020making} as $$D_{LCA}(y', y) := f(y)-f(N_{LCA}(y,y'))$$
where $f(y) \ge f(N_{LCA}(y,y'))$ and $N_{LCA}(y',y)$ denotes the lowest common ancestor class node for classes $y$ and $y'$ within the hierarchy, and $f(\cdot)$ represents a function of a node, such as the tree depth or entropy. We use the information content as described in \cite{valmadre2022hierarchical}.
For each sample $X_i$ in the given dataset $\mathcal{M} := {X_1,\dots,X_n}$: 
$$ D_{LCA}(model, \mathcal{M}) := \frac{1}{n}\sum_{i=1}^n
D_{LCA}(\widehat{y}_i, y_i) \iff y_i \ne \widehat{y}_i$$ 
where $\widehat{y}_i$ is the predicted class for sample $X_i$ using the model, $y_i$ is the ground-truth class for sample $X_i$, and $y_i \ne \widehat{y}_i$. Intuitively, a model with a lower LCA distance demonstrates a greater semantic understanding of class ontology in WordNet.

We can also derive the generalized form of LCA distance to settings where the model outputs a distribution over all possible classes for each sample (like using softmax), please refer to appendix~\ref{ELCA_sup} for details.

\section{Discussion: The Suitability of LCA as a Benchmark for Model Generalization}
\label{section_hypothesis}

This section explores the hypothesis linking LCA distance with a model's generalization ability and discusses how these insights can be meaningfully and actionably applied.

Our primary motivation is to use class hierarchy to capture correlation invariances across training environments, as proposed in the seminal work on `invariant risk minimization'~\cite{arjovsky2019invariant}. Since the class hierarchy remains consistent across both ID and OOD datasets, it can serve as a surrogate measure of the model's invariant features. Models that generalize well to OOD datasets typically learn universal or non-spurious features from the training dataset that are transferable to OOD datasets \cite{makar2022causally}. Such models are more likely to misclassify an ostrich as another bird rather than a lion. These taxonomy-based mispredictions, quantified using the LCA distance, are shown to be a better indicator of a model's OOD performance in this work. 

\begin{figure}[t]
\centering
\includegraphics[width=\columnwidth]{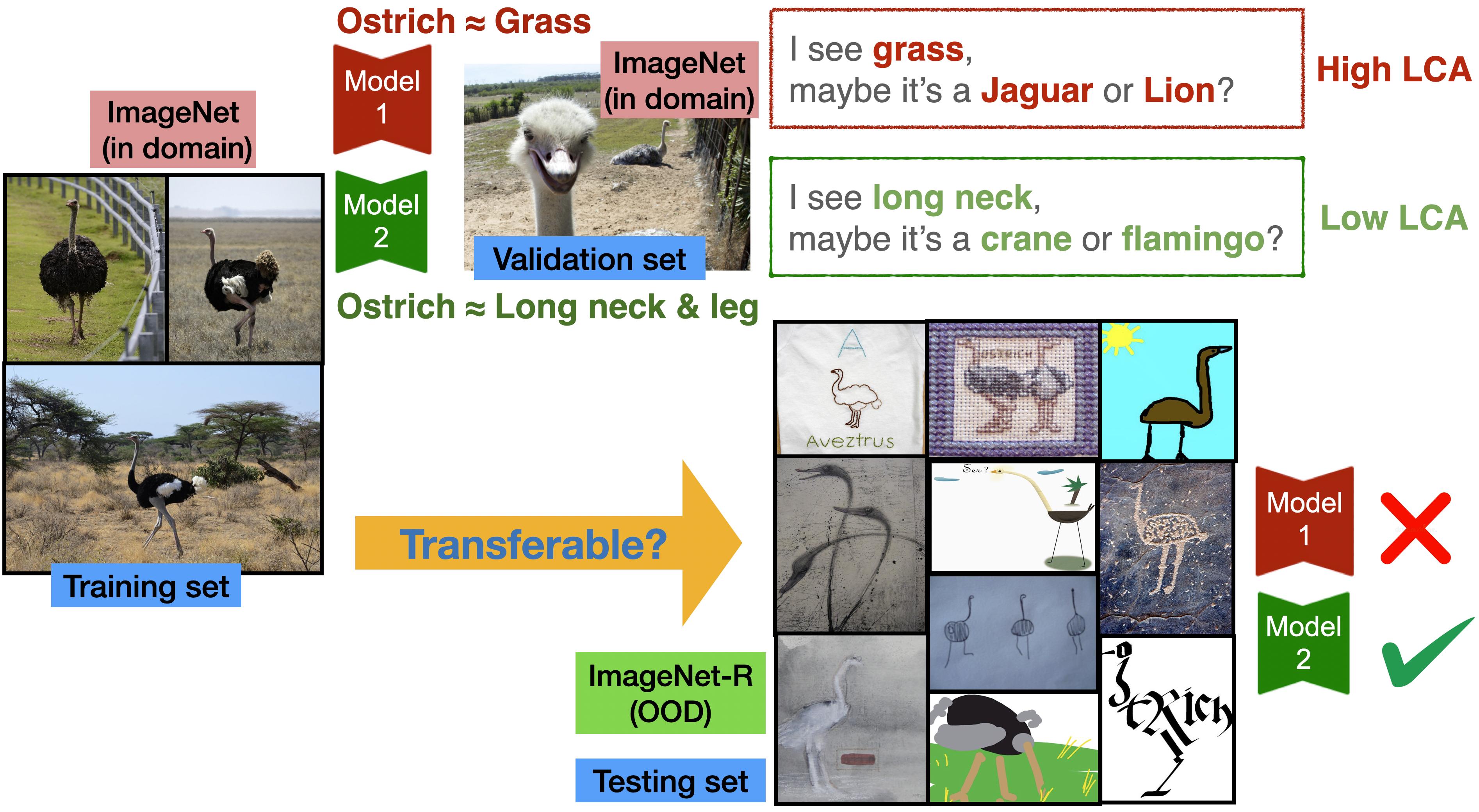}
\vspace{-4mm}
   \caption{\small
\textbf{Capturing transferable features for model generalization.} ImageNet-R maintains shape information~\cite{geirhos2018imagenet} like `long neck', `big belly', and `long legs'. We hypothesize that models with good generalization should capture these transferable features rather than succumbing to spurious correlations such as `grass', thereby tending to predict classes that are semantically closer to the ground-truth. Such models are expected to have low LCA distances between their predictions and the ground-truth.}
\vspace{-3mm}
\label{fig:neck}
\end{figure}

\textbf{Obstacles to Model Generalization.} In deep learning, models often learn predictive features from images by creating discriminative associations to class labels. This approach is susceptible to spurious correlations in the training data \cite{sturm2014simple, torralba2011unbiased, jabri2016revisiting}. For instance, a model might erroneously associate the class `ostriches' with the feature `green grass' in the background, as ostriches often appear in grasslands. These correlations may fail when applied to an OOD dataset that only depicts the semantic concept of `ostriches' \cite{zhang2021understanding}.

\textbf{Essentials for Model Generalization.}
ImageNet-R is a severely shifted OOD dataset where, despite significant distribution shifts, humans can effortlessly identify the correct classes. This is because humans can discern stable features across environments. A model's generalization capability depends on the transferability of the associations learned during training. As benchmarks often simulate human-world ontology, ideally, only features that align with human understanding of object semantics are universally transferable to any constructed OOD dataset. This underscores the importance of identifying transferable features aligning ontology that contribute to robust model generalization.

How can we measure what features a model has learned as predictive during training? The decision-making process of deep neural networks trained end-to-end has become less interpretable. While there have been attempts to decipher this process by forming decision-tree-like models~\cite{wan2020nbdt, gare2022learning} or through learnable activation functions~\cite{liu2024kan}, these efforts have not linked this understanding to measure model generalization.

\textbf{Class Taxonomy Alignment as a Feature Measurement.} 
Class taxonomy or ontology has been widely utilized in literature to indicate class formation~\cite{imagenet_cvpr09,van2018inaturalist} and semantic relationships between classes~\cite{frome2013devise, barz2019hierarchy,wan2020nbdt,redmon2017yolo9000,lin2022continual}, offering a hierarchical organization of classes or categories.

As WordNet encodes class ontology, we hypothesize that transferable features are more likely to be shared among neighboring classes in the hierarchy (e.g., ostrich and crane). In contrast, confounding features are less supported by the hierarchy and tend to appear in less relevant classes that are often more distant in the hierarchy (e.g., lion and ostrich). When a model makes a mistake, its secondary prediction class can provide insight into the predictive features the model has learned during training. Specifically, it reflects that the model perceives the label class and the secondary prediction class to be more similar to each other based on these predictive features.

Consequently, a model that captures more transferable features tends to `make better mistakes~\cite{bertinetto2020making}' by predicting classes that are semantically closer to the ground-truth class. As illustrated in Fig.~\ref{fig:neck}, models that learns to associate ostriches with features like `long legs' and `long neck', which are more transferable to OOD datasets, will likely predict classes like flamingos or cranes. In contrast, a model influenced by spurious correlations and associating ostriches with grass might predict a semantically distant class, like jaguars or lions, which also often appear on grass.

Our method involves measuring model generalization based on the semantic severity of mistakes on in-distribution data. We use the LCA distance, the taxonomic distance between the model's prediction and the ground-truth class in a predefined taxonomic hierarchy like WordNet. If a model consistently makes better mistakes on in-distribution data, we can reasonably assume that the model has captured more transferable features for class discrimination.

\textbf{Class Taxonomy and Mistake Severity.} 
The severity of a mistake in many studies is quantified as the shortest path from the prediction node to the lowest common ancestor (LCA) node in a predefined class hierarchy. This metric, known as `LCA distance' or `hierarchical error', was used in the early years of the ImageNet challenge~\cite{imagenet_cvpr09}. However, it was largely dismissed as it was widely believed to follow the same ordering as Top 1 accuracy \cite{bertinetto2020making}. We revisit this metric and empirically demonstrate that Top 1 accuracy and LCA distance do not always align when VLMs are involved, challenging the common notion. We also appeal for community's attention to revisit this metric with its potential usage in measuring a model's feature awareness to indicate generalization.

\textbf{Causal/Invariant Representation Learning for OOD Generalization.} Recently, there has been an increase in OOD generalization research towards formulating training and testing distributions with causal structures \cite{arjovsky2019invariant, buhlmann2020invariance, peters2016causal}, where distribution shifts primarily arise from interventions or confounding factors. Building upon this, methods \cite{scholkopf2021toward, shen2022weakly, subramanian2022learning} such as CausalVAE \cite{yang2021causalvae} have been proposed, leveraging learned causal representations to capture the causal relationships underlying the data generation process \cite{kaur2022modeling}, which helps mitigate the distributional shifts caused by interventions.

While the connection between OOD generalization and causal concepts is not entirely novel, previous attempts have focused on the causal structure at the latent or abstract level, lacking both interpretability and transparency. Our method aligns with this growing interest in causal/invariant learning, which aims to capture the invariant latent data generation process \cite{kaur2022modeling}. One should expect a model prediction that better aligns with the data generation process to be more robust under intervention, thus generalizing better. Although it is less feasible to model the data generation process of natural images (ImageNet), we essentially follow the same intuition and hypothesize that the WordNet class hierarchy serves as an approximation of invariant correlations between class concepts across environments \cite{arjovsky2019invariant, santurkar2020breeds}, robust to spurious relations in images or shortcuts in learning \cite{makar2022causally}. WordNet is a widely recognized and effective means of encoding semantic relationships between concepts, making it an appropriate proxy for aligning human semantic knowledge \cite{miller1990introduction}. Unlike previous work, WordNet hierarchy provides interpretability, adding a level of transparency to our understanding of model generalization.



\textbf{LCA Illustration with Simulated Data.}
To illustrate our hypothesis that LCA distance can identify features supported by hierarchy, we created a controlled example using a simulated dataset, detailed in Appendix \ref{lca_illustration_full}. In this example, the data generation process is fully controlled. We designed a feature space that includes: 1) transferable causal features supported by hierarchy, 2) non-transferable confounding features not supported by hierarchy, and 3) random noise. Two logistic regression models were trained to mimic models capturing different predictive variables from the training data: one relying on the causal features and the other on the confounding features. The simulation results indicated that the model using causal features supported by hierarchy, which exhibited lower LCA distance, had better out-of-distribution (OOD) accuracy on the in-distribution (ID) test set, despite the model using confounding features achieving better ID accuracy. This example suggests that LCA can effectively identify models that capture relationships aligned with the hierarchical structure. Details in \href{https://github.com/ElvishElvis/LCA-on-the-line}{code snippet}. 

\section{Experiments}
We present experiments benchmarking the relationship between 
Lowest Common Ancestor (LCA) and generalization.

\paragraph{Dataset Setup.}
We leverage 75 pretrained models sourced from open repositories on GitHub for empirical analysis. Our selection comprises 36 Vision Models (VMs) pretrained on ImageNet and supervised from class labels, alongside 39 Vision-Language Models (VLMs) that incorporate language as part of the supervision. 
A comprehensive list of model details, ensuring reproducibility, is provided in Appendix~\ref{section_model_architectures}. We use \textit{ImageNet} \cite{imagenet_cvpr09} as the source in-distribution (ID) dataset, while \textit{ImageNet-v2} \cite{recht2019imagenet}, \textit{ImageNet-Sketch} \cite{hendrycks2019benchmarking}, \textit{ImageNet-Rendition} \cite{hendrycks2021many}, \textit{ImageNet-Adversarial} \cite{hendrycks2021many}, and \textit{ObjectNets} \cite{barbu2019objectnet} are employed as out-of-distribution datasets, exemplifying severe natural distribution shifts. The ImageNet hierarchy, as depicted in \cite{bertinetto2020making}, is utilized. 

Although \textit{ImageNet-v2} is predominantly deemed an OOD dataset in most prior literature~\cite{shankar2020evaluating, miller2021accuracy,baek2022agreement}, our experiments suggest that \textit{ImageNet-v2} aligns more closely with ImageNet than other OOD datasets; we delve into these details in Appendix~\ref{section_discussion}.

Note that the terms in-distribution (ID) and out-of-distribution (OOD) are not model-specific in this context. Due to the varying distribution of training data across different models, ImageNet may not necessarily represent ID data for models like CLIP, where the training data distribution is not explicitly known. Instead, ID and OOD are relative concepts. ImageNet is used as a reference anchor dataset, serving as a baseline to evaluate the generalization capabilities of models on OOD datasets. This approach aligns with prior work, allowing us to consistently measure the shift in performance from ID to OOD datasets, despite the differences in the training data distributions of the models.

\paragraph{Metric Setup.} For our correlation experiment, we use \textit{\textbf{$R^2$} (Coefficient of Determination)} and \textit{PEA (Pearson correlation coefficient)} to measure the strength and direction of linear relationships between two variables. Additionally, we employ \textit{KEN (Kendall rank correlation coefficient)} and \textit{SPE (Spearman rank-order correlation coefficient)} to assess the correspondence of the rankings of two variables.

The importance of these measurements lies in their different focuses. \textbf{Linearity measures}, such as $R^2$ and PEA, are primarily concerned with the fit of a linear model to data points, allowing us to quantify the predictability of changes in one variable based on the other. \textbf{Ranking measures}, like KEN and SPE, provide insights into how the rankings of variables relate to each other, which is crucial in downstream applications such as image retrievals and search engine optimization, where understanding and predicting the ordering of data points is often more important than predicting their exact values. For prediction experiments, we utilize MAE (Mean Absolute Error) to quantify the absolute difference between predictions and ground-truth.

\begin{table}[t]
\small
\centering
\resizebox{\columnwidth}{!}{
\begin{tabular}{lccccccc}
\hline
\textbf{Model} &
  \multicolumn{2}{c}{\textbf{ImgN}} &
  \multicolumn{1}{c}{\textbf{ImgN-v2}} &
  \multicolumn{1}{c}{\textbf{ImgN-S}} &
  \multicolumn{1}{c}{\textbf{ImgN-R}} &
  \multicolumn{1}{c}{\textbf{ImgN-A}} &
  \multicolumn{1}{c}{\textbf{ObjNet}} \\ 
  \cline{1-8} 
 &
  LCA $\downarrow$ &
  \multicolumn{1}{c|}{Top1 $\uparrow$} &
  \multicolumn{1}{c|}{Top1 $\uparrow$} &
  \multicolumn{1}{c|}{Top1 $\uparrow$} &
  \multicolumn{1}{c|}{Top1 $\uparrow$} &
  \multicolumn{1}{c|}{Top1 $\uparrow$} &
  Top1 $\uparrow$
  \\ 
  \cline{2-8} 
ResNet18 &
  6.643 &
  \multicolumn{1}{c|}{0.698} &
  \multicolumn{1}{c|}{0.573} &
  \multicolumn{1}{c|}{0.202} &
  \multicolumn{1}{c|}{0.330} &
  \multicolumn{1}{c|}{0.011} &
  0.272 \\
ResNet50 &
  6.539 &
  \multicolumn{1}{c|}{\textbf{0.733}} &
  \multicolumn{1}{c|}{\textbf{0.610}} &
  \multicolumn{1}{c|}{0.235} &
  \multicolumn{1}{c|}{0.361} &
  \multicolumn{1}{c|}{0.018} &
  0.316 \\ 
CLIP\_RN50 &
  6.327 &
  \multicolumn{1}{c|}{0.579} &
  \multicolumn{1}{c|}{0.511} &
  \multicolumn{1}{c|}{0.332} &
  \multicolumn{1}{c|}{0.562} &
  \multicolumn{1}{c|}{0.218} &
  0.398 \\
CLIP\_RN50x4 &
  \textbf{6.166} &
  \multicolumn{1}{c|}{0.641} &
  \multicolumn{1}{c|}{0.573} &
  \multicolumn{1}{c|}{\textbf{0.415}} &
  \multicolumn{1}{c|}{\textbf{0.681}} &
  \multicolumn{1}{c|}{\textbf{0.384}} &
  \textbf{0.504} 
  \\ \hline
\end{tabular}%
}
\vspace{-2mm}
\caption{\small
\textbf{Model performance corresponds to mistake severity.} 
Results are measured by LCA $\downarrow$ and  Top1 $\uparrow$, respectively.
indicate measurements on a given dataset. We present model comparisons across VMs and VLMs families. In-distribution LCA distance indicate severely shifted OOD performance (ImageNet-S/R/A/O) better than in-distribution (ImageNet) Top1 accuracy (except for ImageNet-v2). \textit{Full 75 models evaluation in Table~\ref{tab:correlation_all}.}}
\label{tab:OOD_LCA}
\end{table}

\subsection{LCA-on-the-Line: In-Distribution Taxonomic Distance (LCA) as an Out-of-Distribution (OOD) Performance Predictor}

\begin{figure*}[ht]
\centering
\resizebox{\textwidth}{!}{
\includegraphics[width=20cm]{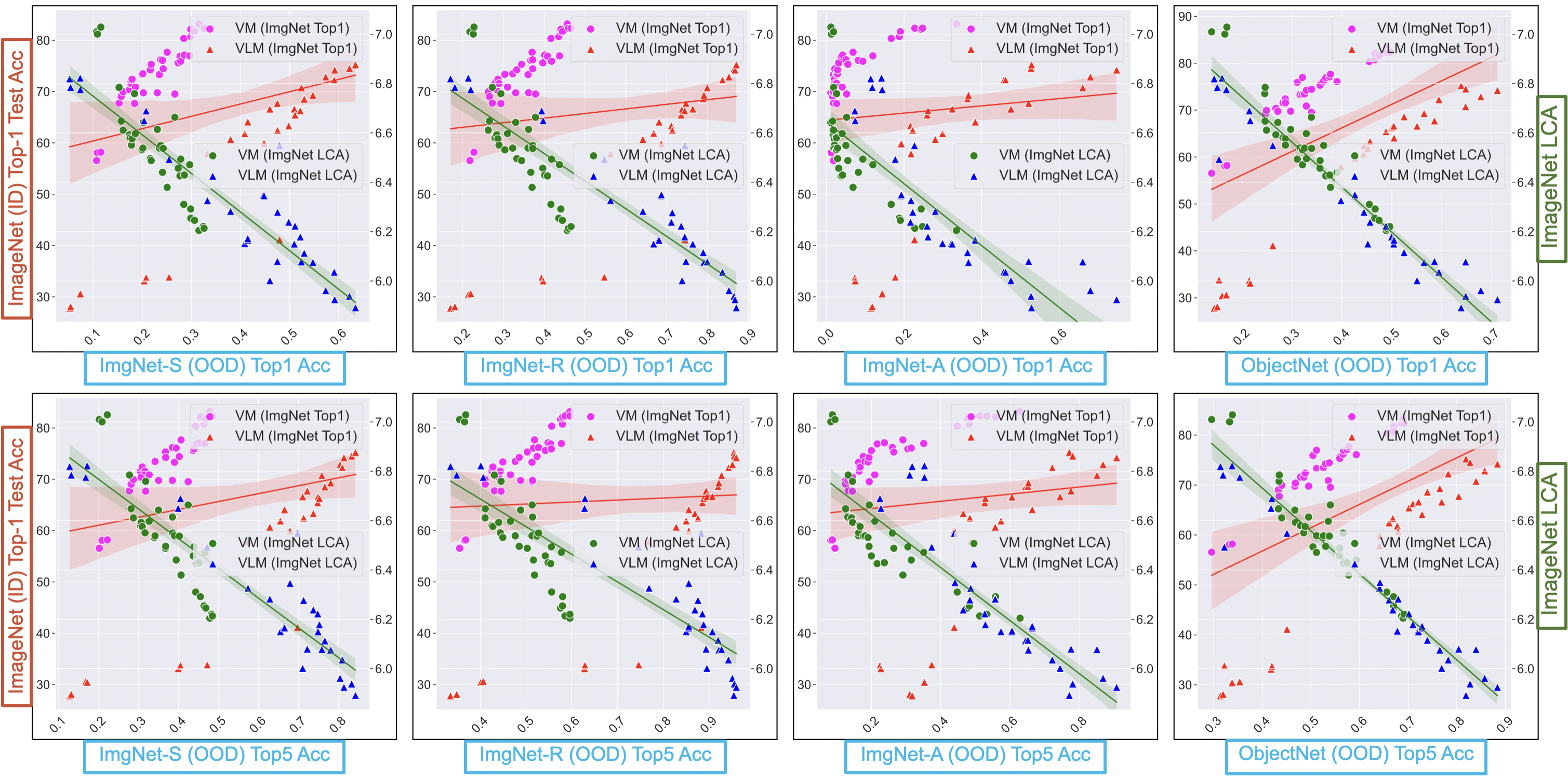}
}
\vspace{-4mm}
\caption{\small
\textbf{Correlating OOD Top-1/Top-5 accuracy (VM+VLM, 75 models) on 4 ImageNet-OOD datasets} visualizing Table~\ref{tab:correlation_all}. 
The plots clearly demonstrate that the \textcolor{LimeGreen}{\textbf{in-distribution LCA distance}} has a stronger correlation with the model's OOD performance across all OOD datasets than \textcolor{Red}{\textbf{accuracy-on-the-line}}~\cite{miller2021accuracy}. Each plot's \textcolor{cyan}{x-axis} represents the OOD dataset metric (with OOD Top-1 in the top row, and OOD Top-5 accuracy in the bottom row) and y-axis represents ImageNet ID test Top-1 accuracy (\textcolor{Red}{left}) and LCA (\textcolor{LimeGreen}{right}); 
\textcolor{Red}{\textbf{Red line}} (\textcolor{Magenta}{Pink dots: VMs} and \textcolor{Red}{Red dots: VLMs}) represents in-distribution classification accuracy (Top-1); \textcolor{LimeGreen}{\textbf{Green line}} (\textcolor{LimeGreen}{Green dots: VMs} and \textcolor{Blue}{Blue dots: VLMs}) denotes in-distribution taxonomic distance (LCA). As interpreted in Figure~\ref{fig:explain_vlm}, accuracy-on-the-line only explains generalization of models within similar settings (VMs or VLMs), but does not unify both settings.}
\label{fig:indicator}
\end{figure*}

Accuracy-on-the-line~\cite{miller2021accuracy} corroborated that a model's in-distribution (ID) accuracy and its out-of-distribution (OOD) accuracy are largely considered to be strongly correlated. This potent correlation forms a significant baseline for comparison in our research. Unlike the framework presented in~\cite{miller2021accuracy}, which only compares models within the same modality, our work bridges the gap by contrasting models of different modalities, involving both Vision Models (VM) and Vision-Language Models (VLM). In addition to the Top1 OOD accuracy, we also incorporate Top5 OOD accuracy, yielding a more comprehensive evaluation of model generalization.

As displayed in Table~\ref{tab:OOD_LCA} and \ref{tab:correlation_all}, the ImageNet in-distribution accuracy~\cite{miller2021accuracy} forms a robust predictor for most OOD datasets, when the comparison is limited to models with similar setups (VMs or VLMs). However, this predictor fails to provide a unified explanation of generalization across models from both families. 
As highlighted in Figure~\ref{fig:indicator} (indicated in \textcolor{Red}{red line}), when adhering to Accuracy-on-the-Line'~\cite{miller2021accuracy}, all four OOD datasets plotted showcase two separate linear trends, representing models that belong to each family. This observation aligns with ~\cite{cherti2022reproducible}, where it was found that VLM models, despite exhibiting significantly lower ID accuracy, could attain higher OOD performance than their state-of-the-art VM counterparts.

As shown in Figure~\ref{fig:explain_vlm}, our method, adopting in-distribution LCA distance, could unify models from both families.
As demonstrated in Table~\ref{tab:correlation_all} and Figure~\ref{fig:indicator} (colored in \textcolor{LimeGreen}{green line}), the severity of in-distribution mistakes serves as a more effective indicator of model performance than in-distribution accuracy. It consistently exhibits a strong linear correlation with all OOD benchmark accuracies for natural distribution shifts (both $R^2$ and the Pearson correlation coefficient exceed 0.7, while~\cite{miller2021accuracy} drop to 0 in ImageNet-A). Notably, our experiments showed that \cite{miller2021accuracy} is a more reliable indicator solely for ImageNet-v2, given its visual similarity to ImageNet. We will further discuss this in Appendix \ref{section_discussion}. 

Our method restores the ``on-the-line'' linear relationship in front of both VMs and VLMs. Our method provides a compelling alternative to understand why vision-language models with lower in-distribution accuracy might generalize better to OOD datasets than vision models.

\begin{table}[t]
\small
\centering
\resizebox{\columnwidth}{!}{
\begin{tabular}{llllllllllll}
\hline
  \multicolumn{2}{c}{\textbf{Element}} &
  \multicolumn{2}{c}{\textbf{ImgN-v2}} &
  \multicolumn{2}{c}{\textbf{ImgN-S}} &
  \multicolumn{2}{c}{\textbf{ImgN-R}} &
  \multicolumn{2}{c}{\textbf{ImgN-A}} &
  \multicolumn{2}{c}{\textbf{ObjNet}} \\ 

\cline{1-2} \cline{3-4} 
\cline{5-6} \cline{7-8} \cline{9-10} \cline{11-12} 

  ID &
  OOD &
  $R^2$ &
  PEA &
  $R^2$ &
  PEA &
  $R^2$ &
  PEA &
  $R^2$ &
  PEA &
  $R^2$ &
  PEA \\ \hline

  Top1 &
  Top1 &
  \textbf{0.962} &
  \textbf{0.980} &
  0.075 &
  0.275 &
  0.020 &
  0.140 &
  0.009 &
  0.094 &
  0.273 &
  0.522 \\

  LCA &
  Top1 &
  0.339 &
  0.582 &
  \textbf{0.816} &
  \textbf{0.903} &
  \textbf{0.779} &
  \textbf{0.883} &
  \textbf{0.704} &
  \textbf{0.839} &
  \textbf{0.915} &
  \textbf{0.956} \\ \cline{1-12} 

  Top1 &
  Top5 &
  \textbf{0.889} &
  \textbf{0.943} &
  0.052 &
  0.229 &
  0.004 &
  0.060 &
  0.013 &
  0.115 &
  0.262 &
  0.512 \\

  LCA &
  Top5 &
  0.445 &
  0.667 &
  \textbf{0.811} &
  \textbf{0.901} &
  \textbf{0.738} &
  \textbf{0.859} &
  \textbf{0.799} &
  \textbf{0.894} &
  \textbf{0.924} &
  \textbf{0.961} \\ \hline
\end{tabular}
}
\vspace{-3mm}
\caption{\small
\textbf{Correlation measurement by $R^2$ and $PEA$
of ID LCA/Top1 with OOD Top1/Top5 across 75 models} (36 VMs and 39 VLMs) as shown in Figure~\ref{fig:indicator}.
We demonstrate that LCA has a strong correlation with OOD performance on all listed datasets (except ImageNet-v2). We take the absolute value of all correlations for simplicity. \textit{Full table containing results of VMs-only and VLMs-only in Table \ref{tab:correlation_all_full}. Measurements from the KEN and SPE show a similar trend as seen in Section \ref{section_supplementary_result}.}
}
\label{tab:correlation_all}
\end{table}

\begin{table}[t]
\small
\centering
\resizebox{\columnwidth}{!}{
\begin{tabular}{llllllllllll}
\hline
   Methods             & \textbf{ImgN-v2} & \textbf{ImgN-S} & \textbf{ImgN-R} & \textbf{ImgN-A} & \textbf{ObjNet} \\ \hline
  ID Top1~\cite{miller2021accuracy}       & \textbf{0.040} & 0.230          & 0.277          & 0.192          & 0.178          \\
     AC~\cite{hendrycks2016baseline}     &  \underline{0.043}  & \underline{0.124}          & \textbf{0.113}          & 0.324          & \underline{0.127}          \\
     Aline-D~\cite{baek2022agreement}       & 0.121          & 0.270          & 0.167          & 0.409          & 0.265          \\
     Aline-S~\cite{baek2022agreement}       & 0.072          & 0.143          & 0.201          & \underline{0.165}          & 0.131          \\
  (Ours) ID LCA & 0.162               & \textbf{0.093}      & \underline{0.114}      & \textbf{0.103}      & \textbf{0.048}     \\ 
  \hline
\end{tabular}
}
\vspace{-3mm}
\caption{\small
\textbf{Error prediction of OOD datasets} across 75 models of diverse settings measured by MAE loss  $\downarrow$.
We mark the best and second best method \textbf{bold} and  \underline{underline}, respectively.
Despite ImageNet (ID) accuracy remaining a significant indicator of ImageNet-v2 (OOD) accuracy, 
the ID LCA serves as a more robust error predictor across the four diverse OOD datasets. \textit{Refer to Table \ref{tab:errorPredict_full} for detailed results of VMs-only and VLMs-only.}
}
\label{tab:errorPredict}
\end{table}

\subsection{Predicting OOD Performance via ID LCA}
We further highlight the effectiveness of the LCA-on-the-Line by estimating model OOD performance using a linear function derived from in-distribution LCA distance. For comparison, we included four competitive baselines: \textit{Average Confidence (AC)}, which leverages OOD logits after temperature scaling; two methods from \textit{Agreement-on-the-Line (Aline-D and Aline-S)}, utilizing consensus of pairs of models on OOD benchmarks; and \textit{`Accuracy on the Line' (ID Top1)}, employing in-distribution accuracy of established measurement models to fit a linear function. Instead of performing a probit transform as done in ~\cite{baek2022agreement} and ~\cite{miller2021accuracy}, we implemented min-max scaling because LCA does not fall within the [0,1] range.


As illustrated in Table \ref{tab:errorPredict}, 
in-distribution LCA distance proves to be a significantly more robust OOD error predictor than other baselines across four OOD benchmarks with varying distribution shifts. This robustness is especially evident for ImageNet-A, an adversarial dataset derived from ResNet50's misclassifications on ImageNet. Consequently, models pre-trained on ImageNet tend to underperform on this dataset, especially those with lower accuracy than ResNet50. This leads to decreased robustness for in-distribution indicators like in-distribution accuracy~\cite{miller2021accuracy}, methods calibrated from in-distribution validation sets~\cite{hendrycks2016baseline}, and OOD agreement of models from different families~\cite{baek2022agreement}. In contrast, LCA, which relies solely on the relative ranking of class predictions from a single model, is less sensitive to these issues and thus delivers more consistent performance. This further underscores the efficacy of LCA as a powerful predictor in challenging OOD scenarios.

\subsection{Enhancing Generalization via  Taxonomy Alignment}

Building upon the earlier discussion, we explore how the devised method can be utilized to enhance a model's generalization capability.

\subsubsection{Inferring Class Taxonomy from a Pretrained Model via K-Means Clustering}

In the previous experiment, we adopted the WordNet hierarchy as class taxonomy to calculate LCA distance. While the number of publicly available datasets providing class taxonomy is limited \cite{imagenet_cvpr09, van2018inaturalist}, the usefulness of our method is unquestionable. Hence, we propose a method to construct a latent class taxonomy given a well-trained model on the task, expanding the potential applications of our work. We show that such a constructed taxonomy could achieve similar correlational performance to the WordNet hierarchy.

\begin{figure}[t]
\centering
\includegraphics[width=\columnwidth]{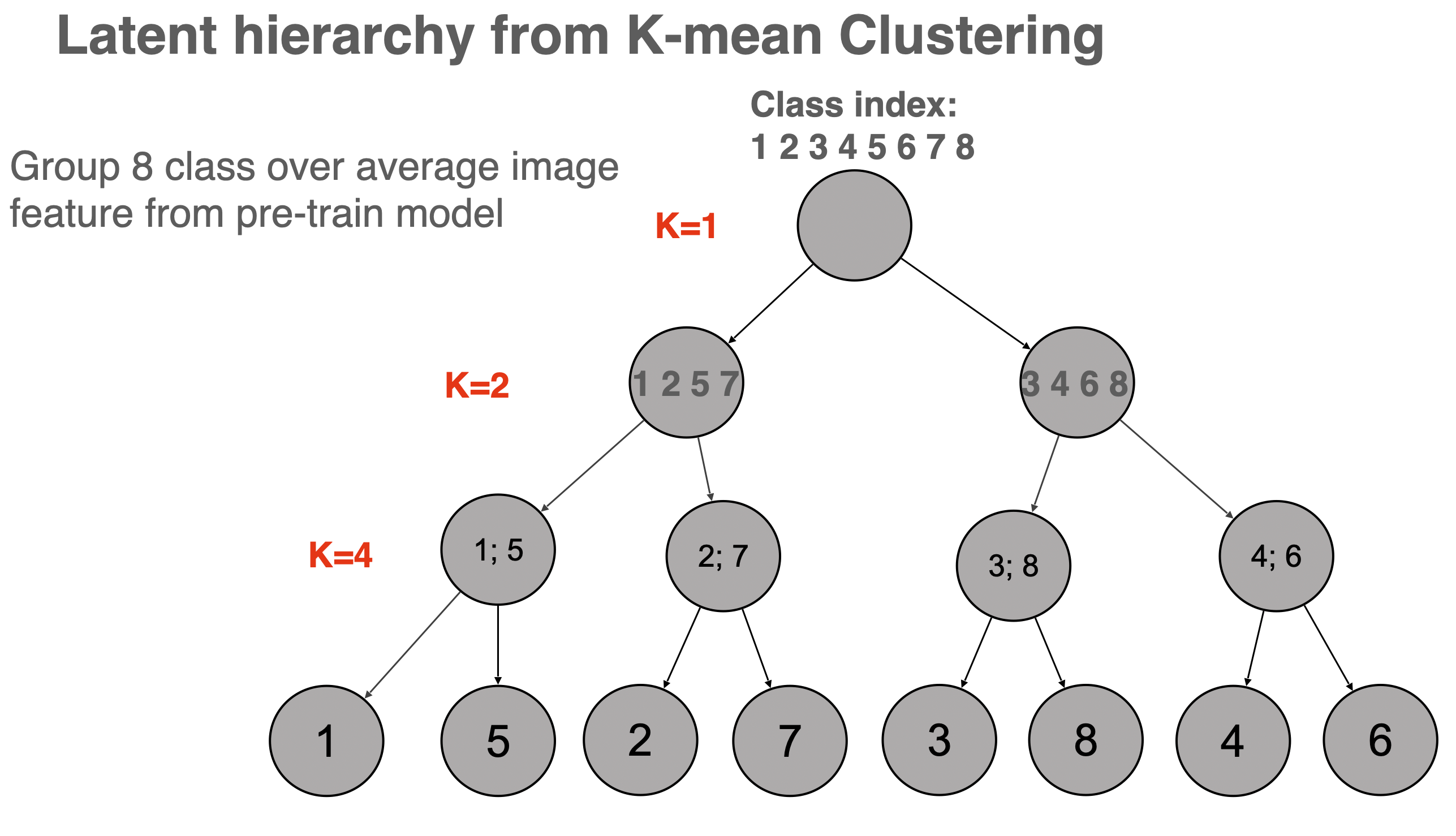}
\vspace{-7.5mm}
\caption{\small
\textbf{Hierarchical structure of image feature clustering using K-means.} 
We construct latent hierarchy through K-means clustering on image features extracted from a pre-trained model. K=1 represent the most generalized cluster, then we incrementally increase the granularity by splitting into K=2 and K=4 clusters. Each node in the hierarchy represents a cluster with the number indicating the class indexes assigned to that cluster. Table~\ref{tab:75_latent_LCA} show that robust performance can be achieved among 75 latent hierarchy constructed from different pretrained models using clustering.}
\vspace{-2mm}
\label{fig:kmean}
\end{figure}

\begin{table}[t]
\centering
\small
\resizebox{\columnwidth}{!}{%
\begin{tabular}{ccccccccl}
\cline{1-8}
\multirow{2}{*}{} &
  \multicolumn{2}{c}{Element} &
  \multirow{2}{*}{ImgN-v2} &
  \multirow{2}{*}{ImgN-S} &
  \multirow{2}{*}{ImgN-R} &
  \multirow{2}{*}{ImgN-A} &
  \multirow{2}{*}{ObjNet} &
   \\ \cline{2-3}
                & ID   & OOD  &                &                &                &                &                &   \\ \cline{1-8}
Baseline      & Top1 & Top1 & \textbf{0.980} & 0.275          & 0.140          & 0.094          & 0.522          &   \\
WordNet      & LCA & Top1 & 0.582 & \textbf{0.903}          & \textbf{0.883}         & \textbf{0.839}          & \textbf{0.956}          &   \\
\hline
\multicolumn{8}{l}{LCA (Statistical Measurements calculated from 75 different Latent Hierarchies)}      \\
\hline
Mean & LCA  & Top1 & 0.815          & \textbf{0.773} & \textbf{0.712} & \textbf{0.662} & \textbf{0.930} &   \\
Min  & LCA  & Top1 & 0.721          & 0.715          & 0.646          & 0.577          & 0.890          &  \\
Max  & LCA  & Top1 & 0.863          & 0.829          & 0.780          & 0.717          & 0.952          &  \\
Std  & LCA  & Top1 & 0.028          & 0.022          & 0.027          & 0.025          & 0.010          &    \\ \cline{1-8}

\end{tabular}%
}
\caption{\small
\textbf{Correlation measurement ($PEA$)
between ID LCA/Top1 and OOD Top1 across 75 latent hierarchies derived from K-means}. Our latent hierarchy construction is robust across 75 different source pretrained models: For each source model, we extracted average class features and applied K-means clustering to construct a latent hierarchy.  We then calculated the LCA distance based on each hierarchy, and aggregated the statistical metric of the 75 groups' Pearson correlation
coefficient ($PEA$) to OOD performance (essentially 75 groups of data from Table~\ref{tab:correlation_all}). \emph{We observe that LCA reliably tracks OOD performance even when using different class taxonomies.}}
\label{tab:75_latent_LCA}
\end{table}

The essence of class taxonomy lies in its representation of inter-class distance, encoding class proximity, and identifying which classes cluster closely in feature space. In this spirit, we can construct a class taxonomy matrix using K-means clustering on image features. 
As illustrated in Fig.~\ref{fig:kmean}, 
for the ImageNet dataset,
we adopt a well-trained model as the source pretrained model and extract average class features to cluster data hierarchically at different levels (we use n=9 for the 1000-class ImageNet dataset, as $2^{9}$ < 1000), with an increasing number of clusters to indicate class adjacency. K-mean is performed on each level of hierarchy independently. Experiments in Table~\ref{tab:75_latent_LCA} show that our method is very robust regardless of which model was used as the source model to construct the class hierarchy. This result demonstrate the potential in practice to use a latent hierarchy constructed by only one well-trained model for evaluating all models on a given task. Further implementation details are provided in Appendix~\ref{section_kmean_apx}.

\begin{table*}[t]
\centering
\resizebox{\textwidth}{!}{%
\begin{tabular}{lcccccccccccc}
\hline
  Hierarchy Source: WordNet&
  \multicolumn{2}{c}{ImgNet} &
  \multicolumn{2}{c}{ImgNet-V2} &
  \multicolumn{2}{c}{ImgNet-S} &
  \multicolumn{2}{c}{ImgNet-R} &
  \multicolumn{2}{c}{ImgNet-A} &
  \multicolumn{2}{c}{ObjectNet} \\ \cline{1-13} 
\textbf{  Backbone Models}
    &
  Baseline &
  \textbf{Ours} &
  Baseline &
  \textbf{Ours} &
  Baseline &
  \textbf{Ours} &
  Baseline &
  \textbf{Ours} &
  Baseline &
  \textbf{Ours} &
  Baseline &
  Ours \\ \hline
ResNet 18~\cite{he2016deep} &
  69.4 &
  \textbf{69.4 \textcolor{Gray}{(+0.0)}} &
  56.4 &
  \textbf{56.9 \textcolor{LimeGreen}{(+0.5)}} &
  19.7 &
  \textbf{20.7 \textcolor{LimeGreen}{(+1.0)}} &
  31.9 &
  \textbf{33.8 \textcolor{LimeGreen}{(+1.8)}} &
  1.1 &
  \textbf{1.2 \textcolor{LimeGreen}{(+0.1)}} &
  27.0 &
  \textbf{28.0 \textcolor{LimeGreen}{(+1.0)}} \\
ResNet 50~\cite{he2016deep} &
  79.5 &
  \textbf{79.8 \textcolor{LimeGreen}{(+0.3)}} &
  67.9 &
  \textbf{68.6 \textcolor{LimeGreen}{(+0.7)}} &
  25.5 &
  \textbf{27.7 \textcolor{LimeGreen}{(+2.2)}} &
  36.5 &
  \textbf{42.5 \textcolor{LimeGreen}{(+6.0)}} &
  10.3 &
  \textbf{16.2 \textcolor{LimeGreen}{(+5.9)}} &
  43.2 &
  \textbf{45.5 \textcolor{LimeGreen}{(+2.3)}} \\
VIT-B~\cite{dosovitskiy2020image} &
  75.8 &
  \textbf{75.9 \textcolor{LimeGreen}{(+0.1)}} &
  62.9 &
  \textbf{62.8 \textcolor{BrickRed}{(-0.1)}} &
  27.0 &
  \textbf{27.6 \textcolor{LimeGreen}{(+0.6)}} &
  40.5 &
  \textbf{41.5 \textcolor{LimeGreen}{(+1.0)}} &
  8.0 &
  \textbf{8.6 \textcolor{LimeGreen}{(+0.6)}} &
  27.6 &
  \textbf{28.1 \textcolor{LimeGreen}{(+0.5)}} \\
VIT-L~\cite{dosovitskiy2020image} &
  76.8 &
  \textbf{76.8 \textcolor{Gray}{(+0.0)}} &
  63.9 &
  \textbf{63.8 \textcolor{BrickRed}{(-0.1)}} &
  28.4 &
  \textbf{29.2 \textcolor{LimeGreen}{(+0.8)}} &
  42.2 &
  \textbf{43.6 \textcolor{LimeGreen}{(+1.4)}} &
  10.6 &
  \textbf{11.5 \textcolor{LimeGreen}{(+0.9)}} &
  28.7 &
  \textbf{29.0 \textcolor{LimeGreen}{(+0.3)}} \\
ConvNext~\cite{liu2022convnet} &
  82.0 &
  \textbf{82.1 \textcolor{LimeGreen}{(+0.1)}} &
  70.6 &
  \textbf{71.0 \textcolor{LimeGreen}{(+0.4)}} &
  28.7 &
  \textbf{30.0 \textcolor{LimeGreen}{(+1.3)}} &
  42.4 &
  \textbf{44.3 \textcolor{LimeGreen}{(+1.9)}} &
  21.8 &
  \textbf{25.3 \textcolor{LimeGreen}{(+3.5)}} &
  44.4 &
  \textbf{45.5 \textcolor{LimeGreen}{(+1.1)}} \\
Swin Transformer~\cite{liu2021swin} &
  83.1 &
  \textbf{83.2 \textcolor{LimeGreen}{(+0.1)}} &
  72.0 &
  \textbf{71.9 \textcolor{BrickRed}{(-0.1)}} &
  30.3 &
  \textbf{31.4 \textcolor{LimeGreen}{(+1.1)}} &
  43.5 &
  \textbf{45.3 \textcolor{LimeGreen}{(+1.8)}} &
  29.5 &
  \textbf{32.7 \textcolor{LimeGreen}{(+3.2)}} &
  48.3 &
  \textbf{49.5 \textcolor{LimeGreen}{(+1.2)}} \\ 
  \hline
\end{tabular}%
}
\vspace{-2mm}
\caption{\small
\textbf{Soft labeling with WordNet for Linear Probing}. Baseline: Trained with Cross Entropy only; Ours: Trained with Cross Entropy + LCA soft loss + weight linear interpolation of (CE, CE + soft loss)~\cite{wortsman2022robust}. Results show that integrating soft loss consistently improves model OOD performance, without compromising ID accuracy. Note that in Table~\ref{tab:ablation} of ablation study in pro-OOD setting, we demonstrate that it's possible to further enhance OOD performance at the cost of a slight ID accuracy drop.}
\label{tab:loss}
\end{table*}

\subsubsection{Using Class Taxonomy as Soft Labels}
\label{soft_labels_lca}
In this section, we investigate how leveraging LCA distance can enhance model generalization through improved supervision.
Traditional models maximize the likelihood of the top-1 ground-truth class but often fail to generalize due to overfitting from spurious correlations. 
We argue that a generalizable model should accurately assign likelihoods to all classes in alignment with the class ontology. Building on this insight, we augment the standard cross-entropy loss, which maximizes the top-1 likelihood, with an auxiliary loss that uses soft labels encoded by the normalized pairwise class distance (LCA distance). 
This approach treats the problem as multi-label classification~\cite{lin2022continual}, guiding the model's decision boundary towards a more regularized feature distribution, 
thereby reducing susceptibility to spurious correlations and improving generalization. We balance the contributions of the cross-entropy and auxiliary losses using a lambda term: 
L = $\lambda$ L(CE) + $L(soft_{lca})$. The detailed formulation is provided in Appendix~\ref{linear_setup}.

\textbf{WordNet as Soft Labels.} To evaluate our approach, we trained linear probe layers on five different models using either cross-entropy loss only (Baseline) or 
our cross-entropy plus LCA soft loss.
We compared their performance on six ImageNet test sets.
Inspired by the notion that models exhibit higher confidence where they excel~\cite{wortsman2022robust}, 
we applied linear interpolation between layers trained with cross-entropy and our proposed loss as our final classifier  $W_{\text{interp}} = \alpha W_{ce} + (1 - \alpha) W_{ce+soft}$.
Table~\ref{tab:loss} shows that incorporating LCA soft loss consistently improved OOD performance without compromising ID performance, indicating more regularized decision boundaries beyond training data knowledge. Ablation study is presented in Table~\ref{tab:ablation}.

\textbf{Latent Hierarchy as Soft Labels.} To demonstrate that our method generalizes beyond WordNet hierarchy, we constructed latent hierarchies using K-means clustering on pretrained models, forming soft labels to guide linear probing. We followed the same training procedure as above, using latent hierarchies instead of WordNet to construct soft labels. As shown in Table~\ref{tab:latent_soft}, adopting constructed hierarchies similarly boosted model generalization across all OOD datasets.

\textbf{VLMs Construct Better Soft Labels Compared to VMs.} Drawing on the intuition of model distillation~\cite{hinton2015distilling}, the hierarchy constructed from a model's pretrained features partially encapsulates the model's interpretation of interclass relationships. Thus we also examined if the source model affects the quality of derived soft labels. Figure~\ref{fig:LCA_matrix} visualizes pair-wise LCA distance matrices for ImageNet data using hierarchies from different models. 

We further conducted a correlation study using latent hierarchies generated from all 75 pretrained models, comparing the source model's ID LCA evaluated on WordNet, with generalization performance from derived soft labels. Table~\ref{tab:lca_distance} reveals a moderate-strong correlation on ImageNet S/R/A, supported by visualizations in Fig.~\ref{fig:lca_distance_matrix_all}. The findings verify that a latent hierarchy derived from a more generalizable model (aligned closer to the WordNet hierarchy) provides higher quality in guiding the linear probe model training to be more generalizable. This visualization also shows that soft labels constructed from VLMs lead to better generalization. Since soft labels are derived from mean class feature clustering, this suggests that VLMs' superior generalization may stem from more regularized feature space distributions over encoded class centroids. Future work should explore the reasons behind VLMs' aligned feature spaces, potentially due to high-level language supervision.

{
\setlength{\tabcolsep}{0.15em} 
\begin{table}[t]
\centering
\resizebox{\columnwidth}{!}{%
\begin{tabular}{ccccccccc}
\hline
\multicolumn{1}{l}{Backcbone Model:ResNet-18} & \multicolumn{2}{c}{ImgNet-S} & \multicolumn{2}{c}{ImgNet-R} & \multicolumn{2}{c}{ImgNet-A} & \multicolumn{2}{c}{ObjectNet} \\ \hline
\textbf{Hierarchy Sources} & Baseline & Interp               & Baseline & Interp               & Baseline & Interp              & Baseline & Interp               \\ \hline
MnasNet                    & 19.7     & 20.2 \textcolor{LimeGreen}{(+0.5)} & 31.9     & 32.4 \textcolor{LimeGreen}{(+0.5)} & 1.1      & 1.7 \textcolor{LimeGreen}{(+0.6)} & 27.0     & 28.1 \textcolor{LimeGreen}{(+1.1)} \\
ResNet 18                  & 19.7     & 20.2 \textcolor{LimeGreen}{(+0.5)} & 31.9     & 32.4 \textcolor{LimeGreen}{(+0.5)} & 1.1      & 1.8 \textcolor{LimeGreen}{(+0.7)} & 27.0     & 28.2 \textcolor{LimeGreen}{(+1.2)} \\
vit-l-14                   & 19.7     & 20.8 \textcolor{LimeGreen}{(+1.2)} & 31.9     & 33.2 \textcolor{LimeGreen}{(+1.3)} & 1.1      & 2.0 \textcolor{LimeGreen}{(+0.9)}  & 27.0     & 28.3 \textcolor{LimeGreen}{(+1.3)} \\
OpenCLIP(vit-l-14)                         & 19.7  & 20.9 \textcolor{LimeGreen}{(+1.3)} & 31.9  & 33.7 \textcolor{LimeGreen}{(+1.8)} & 1.1   & 2.1 \textcolor{LimeGreen}{(+1.0)}   & 27.0  & 28.5 \textcolor{LimeGreen}{(+1.5)}  \\ \hline
WordNet                    & 19.7     & \textbf{21.2 \textcolor{LimeGreen}{(+1.5) }}          & 31.9     & \textbf{35.1 \textcolor{LimeGreen}{(+3.2) }}         & 1.1      & \textbf{1.4 \textcolor{LimeGreen}{(+0.4) }}         & 27.0     & \textbf{28.6 \textcolor{LimeGreen}{(+1.6) }}         \\ \hline
\end{tabular}%
}
\vspace{-3mm}
\caption{\textbf{Soft Labeling with Latent Hierarchies for Linear Probing on ResNet-18.} Instead of using WordNet to construct soft labels in Table~\ref{tab:loss}, we adopted latent hierarchies constructed from pre-trained models using K-means clustering. Results show that using latent hierarchies also delivers a generalization boost compared to the baseline, although it is less significant than using WordNet. Experiments are listed here with the pro-OOD setting in Table~\ref{tab:ablation}.}
\label{tab:latent_soft}
\vspace{-2mm}
\end{table}
}

\subsubsection{Improving Generalization by Class Taxonomy Alignment with Prompt Engineering}
\label{apx:prompt_engg}
In this section, we discuss results on enhancing model generalization through prompt engineering in VLMs.






For VLM, integrating taxonomy-specific knowledge during zero-shot evaluation is straightforward.
The WordNet hierarchy naturally indicates inter-class distances from class definitions. 
For example, `dalmatian' and `husky' are semantically close, both originating from the parent node `dog'.
We detail the results with
CLIP-ViT32~\citep{radford2021learning} in Table~\ref{tab:parent}.
To test our hypothesis, we explicitly integrated hierarchical taxonomy relationships into the prompt for zero-shot VLM predictions. 
The prompt was designed as \textbf{`A, which is a type of B, which is a type of C'}, guiding the model to make taxonomy-aligned predictions. Additionally, we conducted two ablation studies: 1) \textbf{Stack Parent:} providing the correct taxonomy path without informing the model of the class name relationships; 
and 2) \textbf{Shuffle Parent:} informing the model of the hierarchical `is-a' relationship but providing an incorrect taxonomy relationship randomly sampled from the tree. Our results demonstrate that informing the model of both the correct taxonomy and their hierarchical relationships significantly improves generalization. This improvement is evidenced by enhancements in Top-1 accuracy and test-time Cross-Entropy (CE) across all datasets for all tested models.

\section{Conclusions}

This work revitalizes the use of LCA distance, leveraging class taxonomies such as WordNet, to indicate model OOD performance. We assess the severity of model mispredictions in a manner agnostic to model modality, architecture or training data source, establishing a comprehensive metric for evaluating model generalization. Our findings across multiple ImageNet-OOD datasets highlight the superiority of LCA distance in reflecting the generalization capabilities of models trained with either class labels (VMs) or captions (VLMs), surpassing the traditional reliance on in-distribution Top-1 accuracy~\cite{miller2021accuracy}. To extend the application of LCA distance measurement to any dataset, we introduce a method for creating latent hierarchies using K-means clustering, showcasing the resilience of LCA distance regardless of the applied taxonomy or hierarchy. Additionally, we demonstrate that aligning model predictions with class taxonomies, through soft labels or prompt engineering, can enhance model generalization. Our results on demonstrating VLMs' lower LCA distance and better soft label construction offer new insights into VLMs' superior model generalization from a feature distribution perspective.

Future research could focus on providing theoretical justification for the LCA-on-the-Line framework. For instance, exploring causal discovery~\cite{brouillard2020differentiable} methods on the ImageNet dataset to construct a causal graph between classes and underlying variables may offer a more accurate reflection of causal relationships between classes.

\clearpage

\section*{Acknowledgements}
Authors thank Deva Ramanan for insightful discussions, and Hualiang Wang for valuable feedback on the manuscript. The work was partially supported by the CMU Argo Research Center.
Shu Kong is partially supported by the University of Macau (SRG2023-00044-FST).

\section*{Limitation}
While we benchmarked and used LCA based on class hierarchy to measure model generalization, the findings from this work indicate that it is not an effective indicator for datasets visually similar to in-distribution data (like ImageNet2, more discussion in Appendix~\ref{section_discussion}). For these datasets, in-distribution Top1 remains a strong indicator, which potentially limits the utility of LCA. Also, it is expected that LCA will show a weaker discrimination between models on datasets with small number of classes (like CIFAR~\cite{cifar10}).

\section*{Impact Statement}

This research aims to enhance our understanding of model generalization mechanisms. However, it is crucial to recognize its potential misuse, such as in guiding adversarial attacks that reduce the generalization capabilities of existing models. 
Although not the intended purpose of our research, the dual potential of our findings in model generalization underscores the need for robust, 
secure model development and the implementation of ethical guidelines for deploying this knowledge.

\bibliography{example_paper}
\bibliographystyle{icml2024}

\newpage
\appendix
\onecolumn

\section{Model Architectures}
\label{section_model_architectures}
We list all models used in ours experiment as follows, including 36 Vision Only Models (VMs) and 39 Vision-Language Models (VLMs).

\begin{table}[bh!]
\centering
\resizebox{0.8\textwidth}{!}{%
\begin{tabular}{|c|c|c|l|}
\hline
\textbf{Model Category} &
  \textbf{Architecture} &
  \textbf{Number of models} &  \textbf{Checkpoint Link} 
   \\ \hline
\multirow{16}{*}{VM (Vision-Only-Models)} &
  AlexNet~\citep{krizhevsky2017imagenet} &
  1 &
   \href{https://pytorch.org/vision/main/models/generated/torchvision.models.alexnet.html}{alexnet} \\ \cline{2-4} 
 &
  ConvNeXt~\citep{liu2022convnet} &
  1 &
   \href{https://pytorch.org/vision/main/models/generated/torchvision.models.convnext\_tiny.html}{convnext\_tiny} \\ \cline{2-4} 
 &
  DenseNet~\citep{huang2017densely} &
  4 &
  \begin{tabular}[c]{@{}l@{}} \href{https://pytorch.org/vision/main/models/generated/torchvision.models.densenet121.html}{densenet121}\\  \href{https://pytorch.org/vision/main/models/generated/torchvision.models.densenet161.html}{densenet161}\\  \href{https://pytorch.org/vision/main/models/generated/torchvision.models.densenet169.html}{densenet169}\\  \href{https://pytorch.org/vision/main/models/generated/torchvision.models.densenet201.html}{densenet201}\end{tabular} \\ \cline{2-4} 
 &
  EfficientNet~\citep{tan2019efficientnet} &
  1 &
   \href{https://pytorch.org/vision/main/models/generated/torchvision.models.efficientnet\_b0.html}{efficientnet\_b0} \\ \cline{2-4} 
 &
  GoogLeNett~\citep{szegedy2015going}  &
  1 &
   \href{https://pytorch.org/vision/main/models/generated/torchvision.models.googlenet.html}{googlenet} \\ \cline{2-4} 
 &
  Inceptionv3~\citep{szegedy2016rethinking} &
  1 &
   \href{https://pytorch.org/vision/main/models/generated/torchvision.models.quantization.inception\_v3.html}{inceptionV3} \\ \cline{2-4} 
 &
  MnasNet~\citep{tan2019mnasnet}  &
  4 &
  \begin{tabular}[c]{@{}l@{}} \href{https://pytorch.org/vision/main/models/generated/torchvision.models.mnasnet0\_5.html}{mnasnet0.5}\\  \href{https://pytorch.org/vision/main/models/generated/torchvision.models.mnasnet0\_75.html}{mnasnet0.75}\\  \href{https://pytorch.org/vision/main/models/generated/torchvision.models.mnasnet1\_0.html}{mnasnet1.0}\\  \href{https://pytorch.org/vision/main/models/generated/torchvision.models.mnasnet1\_3.html}{mnasnet1.3}\end{tabular} \\ \cline{2-4} 
 &
  Mobilenet-V3~\citep{howard2019searching} &
  2 &
  \begin{tabular}[c]{@{}l@{}} \href{https://pytorch.org/vision/main/models/generated/torchvision.models.mobilenet\_v3\_small.html}{mobilenetv3\_small}\\  \href{https://pytorch.org/vision/main/models/generated/torchvision.models.mobilenet\_v3\_large.html}{mobilenetv3\_large}\end{tabular} \\ \cline{2-4} 
 &
  Regnet~\citep{radosavovic2020designing} &
  1 &
   \href{https://pytorch.org/vision/main/models/generated/torchvision.models.regnet\_y\_1\_6gf.html}{regnet\_y\_1\_6gf} \\ \cline{2-4} 
 &
  Wide ResNet~\citep{zagoruyko2016wide}  &
  1 &
   \href{https://pytorch.org/vision/main/models/generated/torchvision.models.wide\_resnet101\_2.html}{wide\_resnet101\_2} \\ \cline{2-4} 
 &
  ResNet~\citep{he2016deep} &
  5 &
  \begin{tabular}[c]{@{}l@{}} \href{https://pytorch.org/vision/main/models/generated/torchvision.models.resnet18.html}{resnet18}\\  \href{https://pytorch.org/vision/main/models/generated/torchvision.models.resnet34.html}{resnet34}\\  \href{https://pytorch.org/vision/main/models/generated/torchvision.models.resnet50.html}{resnet50}\\  \href{https://pytorch.org/vision/main/models/generated/torchvision.models.resnet101.html}{resnet101}\\  \href{https://pytorch.org/vision/main/models/generated/torchvision.models.resnet152.html}{resnet152}\end{tabular} \\ \cline{2-4} 
 &
  ShuffleNet~\citep{zhang2018shufflenet} &
  1 &
  \href{https://pytorch.org/vision/stable/models/generated/torchvision.models.shufflenet\_v2\_x2\_0.html}{shufflenet\_v2\_x2\_0} \\ \cline{2-4} 
 &
  SqueezeNet~\citep{iandola2016squeezenet} &
  2 &
  \begin{tabular}[c]{@{}l@{}} \href{https://pytorch.org/vision/main/models/generated/torchvision.models.squeezenet1\_0.html}{squeezenet1\_0}\\  \href{https://pytorch.org/vision/main/models/generated/torchvision.models.squeezenet1\_1.html}{squeezenet1\_1}\end{tabular} \\ \cline{2-4} 
 &
  Swin Transformer~\citep{liu2021swin}  &
  1 &
   \href{https://pytorch.org/vision/main/models/generated/torchvision.models.swin\_b.html}{swin\_b} \\ \cline{2-4} 
 &
  VGG~\citep{simonyan2014very} &
  8 &
  \begin{tabular}[c]{@{}l@{}} \href{https://pytorch.org/vision/main/models/generated/torchvision.models.vgg11.html}{vgg11}\\  \href{https://pytorch.org/vision/main/models/generated/torchvision.models.vgg13.html}{vgg13}\\  \href{https://pytorch.org/vision/main/models/generated/torchvision.models.vgg16.html}{vgg16}\\  \href{https://pytorch.org/vision/main/models/generated/torchvision.models.vgg19.html}{vgg19}\\  \href{https://pytorch.org/vision/main/models/generated/torchvision.models.vgg11\_bn.html}{vgg11\_bn}\\  \href{https://pytorch.org/vision/main/models/generated/torchvision.models.vgg13\_bn.html}{vgg13\_bn}\\  \href{https://pytorch.org/vision/main/models/generated/torchvision.models.vgg16\_bn.html}{vgg16\_bn}\\  \href{https://pytorch.org/vision/main/models/generated/torchvision.models.vgg19\_bn.html}{vgg19\_bn}\end{tabular} \\ \cline{2-4} 
 &
  ViT ~\citep{dosovitskiy2020image}&
  2 &
  \begin{tabular}[c]{@{}l@{}} \href{https://pytorch.org/vision/main/models/generated/torchvision.models.vit\_b\_32.html}{vit\_b\_32}\\  \href{https://pytorch.org/vision/main/models/generated/torchvision.models.vit\_l\_32.html}{vit\_l\_32}\end{tabular} \\ \hline
\multirow{4}{*}{VLM (Vision-Language-Models)} &
  ALBEF~\citep{li2021align} &
  1 &
  \href{https://github.com/salesforce/LAVIS/blob/main/lavis/configs/models/albef\_feature\_extractor.yaml}{albef\_feature\_extractor} \\ \cline{2-4} 
 &
  BLIP~\citep{li2022blip} &
  1 &
  \href{https://github.com/salesforce/LAVIS/blob/main/lavis/configs/models/blip\_feature\_extractor\_base.yaml.yaml}{blip\_feature\_extractor\_base} \\ \cline{2-4} 
 &
  CLIP~\citep{radford2021learning} &
  7 &
  \begin{tabular}[c]{@{}l@{}} \href{https://openaipublic.azureedge.net/clip/models/afeb0e10f9e5a86da6080e35cf09123aca3b358a0c3e3b6c78a7b63bc04b6762/RN50.pt}{RN50}\\ \href{https://openaipublic.azureedge.net/clip/models/8fa8567bab74a42d41c5915025a8e4538c3bdbe8804a470a72f30b0d94fab599/RN101.pt}{RN101}\\  \href{https://openaipublic.azureedge.net/clip/models/7e526bd135e493cef0776de27d5f42653e6b4c8bf9e0f653bb11773263205fdd/RN50x4.pt}{RN50x4}\\ \href{https://openaipublic.azureedge.net/clip/models/40d365715913c9da98579312b702a82c18be219cc2a73407c4526f58eba950af/ViT-B-32.pt}{ViT-B-32.pt}\\ \href{https://openaipublic.azureedge.net/clip/models/5806e77cd80f8b59890b7e101eabd078d9fb84e6937f9e85e4ecb61988df416f/ViT-B-16.pt}{ViT-B-16.pt}\\ \href{https://openaipublic.azureedge.net/clip/models/b8cca3fd41ae0c99ba7e8951adf17d267cdb84cd88be6f7c2e0eca1737a03836/ViT-L-14.pt}{ViT-L-14.pt}\\ \href{https://openaipublic.azureedge.net/clip/models/3035c92b350959924f9f00213499208652fc7ea050643e8b385c2dac08641f02/ViT-L-14-336px.pt}{ViT-L-14-336px}\end{tabular} \\ \cline{2-4} 
 &
  OpenCLIP~\citep{cherti2023reproducible} &
  30 &
  \begin{tabular}[c]{@{}l@{}}\href{https://github.com/mlfoundations/open\_clip}{openCLIP}:\\ openCLIP\_('RN101', 'openai')\\ openCLIP\_('RN101', 'yfcc15m')\\ openCLIP\_('RN101-quickgelu', 'openai')\\ openCLIP\_('RN101-quickgelu', 'yfcc15m')\\ openCLIP\_('RN50', 'cc12m')\\ openCLIP\_('RN50', 'openai')\\ openCLIP\_('RN50', 'yfcc15m')\\ openCLIP\_('RN50-quickgelu', 'cc12m')\\ openCLIP\_('RN50-quickgelu', 'openai')\\ openCLIP\_('RN50-quickgelu', 'yfcc15m')\\ openCLIP\_('RN50x16', 'openai')\\ openCLIP\_('RN50x4', 'openai')\\ openCLIP\_('RN50x64', 'openai')\\ openCLIP\_('ViT-B-16', 'laion2b\_s34b\_b88k')\\ openCLIP\_('ViT-B-16', 'laion400m\_e31')\\ openCLIP\_('ViT-B-16', 'laion400m\_e32')\\ openCLIP\_('ViT-B-16-plus-240', 'laion400m\_e31')\\ openCLIP\_('ViT-B-16-plus-240', 'laion400m\_e32')\\ openCLIP\_('ViT-B-32', 'laion2b\_e16')\\ openCLIP\_('ViT-B-32', 'laion2b\_s34b\_b79k')\\ openCLIP\_('ViT-B-32', 'laion400m\_e31')\\ openCLIP\_('ViT-B-32', 'laion400m\_e32')\\ openCLIP\_('ViT-B-32', 'openai')\\ openCLIP\_('ViT-B-32-quickgelu', 'laion400m\_e31')\\ openCLIP\_('ViT-B-32-quickgelu', 'laion400m\_e32')\\ openCLIP\_('ViT-L-14', 'laion2b\_s32b\_b82k')\\ openCLIP\_('ViT-L-14', 'laion400m\_e31')\\ openCLIP\_('ViT-L-14', 'laion400m\_e32')\\ openCLIP\_('coca\_ViT-B-32', 'laion2b\_s13b\_b90k')\\ openCLIP\_('coca\_ViT-L-14', 'laion2b\_s13b\_b90k')\end{tabular} \\ \hline
\end{tabular}%
}
\label{tab:my-table}
\end{table}
\clearpage

\section{Discussion}

\label{section_discussion}

\paragraph{Reestablishing LCA as a Comprehensive Measure of Model Generalization.}

While Top 1 ID accuracy~\citep{miller2021accuracy} demonstrates a clear linear trend with OOD datasets in models with similar training mechanisms, this relationship becomes less distinct across VMs and VLMs. This finding, echoed in earlier studies~\citep{fang2022data,wortsman2022robust,cherti2022reproducible}, suggests a more nuanced understanding of how zero-shot VLMs with lower Top-1 accuracy can outperform competitive vision models in generalizing to unfamiliar datasets. While previous works have emphasized the significant impact of data diversity on generalization~\citep{fang2022data, schuhmann2022laion, kaur2022modeling}, our results indicate that the LCA offers a more all-encompassing assessment of model generalization. 
By considering factors such as training data size, architecture, loss, and others, LCA better measures a model's ability to accurately capture semantic distinctions common across ID and OOD benchmarks. This establishes a comprehensive benchmark that encompasses various generalization factors, addressing the issue of inflated VLM effectiveness on ``Effective Robustness~\citep{taori2020measuring}''. Future research should delve into large-scale analytic studies of generalization factors in conjunction with LCA.

\paragraph{ImageNet-v2 Demonstrates Similar Class Discrimination Features to ImageNet.}
ImageNet-v2, a recollection of ImageNet, is often used as an OOD dataset for ImageNet-based studies~\citep{shankar2020evaluating, miller2021accuracy,baek2022agreement}. Our experiments indicate that ImageNet-v2 more closely resembles ImageNet than other OOD datasets. We hypothesize that the minimal external intervention in ImageNet-v2's data collection process results in visual similarities to ImageNet (as ImageNet-v2 is a recollection of ImageNet), allowing even spurious relationships encoded on ImageNet to transfer successfully to ImageNet-v2. Consequently, models pretrained on ImageNet (VMs) inflate accuracy on ImageNet-v2, disrupting the alignment with trends observed in VLMs.

\paragraph{Is it Possible for a Semantically-Aware (Low LCA) Model to Have Low Top 1 Accuracy?}

Our empirical analysis indicates a correlation: models not specifically tuned on class taxonomy, with lower Top 1 accuracy, tend to exhibit higher LCA distances. However, this relationship is correlational rather than causal. It remains feasible to design a model adversarially so it consistently predicts the semantically nearest class to the true class. In such cases, the model would show a low LCA distance while maintaining zero Top 1 accuracy. Therefore, while a correlation exists between Top 1 accuracy and LCA, causality cannot be inferred, and this relationship can be disrupted under deliberate adversarial training.

\paragraph{Does ImageNet LCA (Taxonomic Distance) Reflect ImageNet Top 1 Accuracy?}

It is often suggested that LCA and Top-1 accuracy exhibit similar trends \textit{on the same dataset}~\citep{imagenet_cvpr09, bertinetto2020making}. Intuitively, a high-performing model better fits the data distribution, leading to fewer severe errors. This pattern generally holds true for models under similar settings (either VM or VLM separately). However, when considering both VM and VLM models, ImageNet and ImageNet-v2 exhibit only a weak correlation between LCA and Top-1 accuracy, whereas other semantically distinct OOD datasets show a stronger relationship (validate in Section~\ref{igmNeta_to_lca_a}). This finding challenges the prevailing belief that in-distribution Top-1 accuracy and LCA maintain the same ranking~\citep{deng2009imagenet,bertinetto2020making}. 

\paragraph{Why do we observe low LCA correlation numbers between IID test sets?}
From previous experiments, we observe that ImageNet LCA (Taxonomic Distance) does not correlate strongly with ImageNet/ImageNet-v2 Top-1 Accuracy, often showing a weak relationship, as illustrated in Figure~\ref{fig:indicator_ID}. We hypothesize that this is due to ID accuracy inflation. In our LCA-on-the-Line framework, LCA is expected to be an unbiased measure of alignment to the class hierarchy. However, the VMs used in this work are trained on ImageNet and tend to     `inflate' ID accuracy when evaluated on IID test sets. As indicated in the bottom right two images of Figure~\ref{fig:indicator_ID}, this inflation might causes datapoints to `shift' in the direction of the red arrow, disrupting the unbiased linear relationship seen in VLMs that were not trained directly on ImageNet. Consequently, we should expect models evaluating LCA on unseen datasets to form a linear relationship, similar to the observed relationship on the other four severely shifted OOD datasets in Figure~\ref{fig:indicator_ID}. Please refer to Section~\ref{igmNeta_to_lca_a} and Table~\ref{tab:correlation_ID} for a numerical comparison.

\section{LCA Illustration with Simulated Data}
\label{lca_illustration_full}


To illustrate the hypotheses in Section \ref{section_hypothesis}: 1) Transferable features are more likely to be supported by the hierarchy and shared among neighboring classes; 2) Confounding features are less supported by the hierarchy and tend to appear in less relevant classes that are often more distant in the hierarchy; 3) LCA is useful in identifying features supported by the hierarchy, we created a simple example using a simulated dataset.

Consider a feature space $\mathbf{x} := (x_1, x_2, x_3) \in \mathbb{R}^3$ and a latent class $z \in {1,2,3,4}$, where class 1 and 2 are similar, and class 3 and 4 are similar. By design, we set the joint distribution of $\mathbf{x}$ and $z$ to follow a mixture of Gaussians, 
where $x_1 \in \{1,3,15,17\}$,
$x_2 \in \{1,17,7,21\}$, $x_3 \in \{0,0,0,0\}$ for each class respectively.

\begin{align}\label{toymodel}
    & \mathbf{x} | z=1 \sim N(\mathbf{\mu}_1, \mathbf{I}), \quad \mathbf{\mu}_1=(1,1,0) \nonumber \\
    & \mathbf{x} | z=2 \sim N(\mathbf{\mu}_2, \mathbf{I}), \quad \mathbf{\mu}_2=(3,17,0) \nonumber\\
    & \mathbf{x} | z=3 \sim N(\mathbf{\mu}_3, \mathbf{I}), \quad \mathbf{\mu}_3=(15,7,0) \nonumber\\
    & \mathbf{x} | z=4 \sim N(\mathbf{\mu}_4, \mathbf{I}), \quad \mathbf{\mu}_4=(17,21,0)
\end{align}

Given a hierarchy preserving class proximity: ${\texttt{root}: (\texttt{class 1, class 2}), (\texttt{class 3, class 4})}$, by design, only feature $x_1$ supports the class hierarchy, as the distance w.r.t $x_1$ between classes 1 \& 2 and classes 3 \& 4 is smaller than those for other pairs. Feature $x_2$ can distinguish all four classes but is not supported by the class hierarchy. Feature $x_3$ is random noise with no predictive power for the latent class.

For the in-distribution (ID) data, all three features are observed, while for the out-of-distribution (OOD) data, only $x_1$ and $x_3$ are observed. From hypothesis in section \ref{section_hypothesis}, $x_1$ can be considered a transferable causal feature because it is supported by the true class hierarchy and is observable in all datasets. In contrast, $x_2$ is a non-transferable confounding feature that does not preserve the class hierarchy and is only observable in the ID data. By design (larger $\mathbf{\mu}$ gap between classes), confounder $x_2$ display stronger discrimination among four classes than $x_1$ on ID data.

We trained two logistic regression models on the in-distribution (ID) dataset, mimicking models that captured different features as predictive variables learned from the training data.
\begin{itemize}
\item Model $f$, which trains on the transferable causal feature $x_1$, and noise feature $x_3$.
\item Model $g$, which trains on the non-transferable confounding feature $x_2$, and noise feature $x_3$.
\end{itemize}

From simulations (10,000 samples across 100 independent trials), we observed the following results listed in Table~\ref{tab:lca_toy}:
\begin{itemize}
\item \textbf{Model $g$} achieved better ID accuracy because it can leverage $x_2$, which distinguishes all four classes effectively in the ID data.
\item \textbf{Model $f$} had better OOD accuracy because $x_1$ is a transferable feature that is also present in the OOD data, supported by the true class hierarchy that's invariant across ID and OOD data. 
\item \textbf{Model $f$} showed better (lower) LCA distance on the ID test set, indicating that it captures the class hierarchy better by relying on the transferable causal feature $x_1$.
\end{itemize}

This example illustrates the hypothesis presented in Section \ref{section_hypothesis} and provides the expected output in Table~\ref{tab:lca_toy}. The results suggest that LCA can effectively identify models that capture relationships aligned with the hierarchical structure. For further details, please refer to \href{https://github.com/ElvishElvis/LCA-on-the-line}{code snippet}. 

\begin{table}[]
\centering
\resizebox{0.7\textwidth}{!}{%
\begin{tabular}{llll}
\hline
                            & ID Top1 Error $\downarrow$ & ID LCA Distance $\downarrow$ & OOD Top1 Error $\downarrow$ \\ 
                            \hline
g(w. confounding feature)   & \ \ \ \textbf{0.1423}        & \ \ \ 2.000           & \ \ \ 0.7503         \\
f(w. transferable feature) & \ \ \ 0.3287        & \ \ \ \textbf{1.005}           & \ \ \ \textbf{0.3197}         \\
Diff                        & \ \textcolor{Red}{+0.1864}      & \ \ \textcolor{LimeGreen}{-0.995}          & \ \ \textcolor{LimeGreen}{-0.4306}        \\ 
\hline
\end{tabular}%
}
\caption{\textbf{Observation from simulation data with 100 trials.} The average ID test accuracy error (i.e. top 1 error)  \contour{red}{\textcolor{red}{\texttt{ID\_Top1\_Error} $\downarrow$}}, ID test LCA distance  \contour{red}{\textcolor{red}{\texttt{ID\_LCA\_Distance} $\downarrow$}}, and OOD test accuracy error \contour{red}{\textcolor{red}{\texttt{OOD\_Top1\_Error} $\downarrow$}} for generalizable ``good'' prediction model $f$ and non-generalizable ``bad'' prediction model $g$ over 100 independent trials. Specifically, we design the data generation process as described in \eqref{toymodel}, and $f$ is ``good'' as it learns to rely on the transferable causal features supported by hiearachy; while $g$ is ``bad'' as it instead relies on the non-transferable confounding features not supported by hiearachy. In this example, ID LCA distance is a better indicator of OOD performance than ID Top1 accuracy, and model f display better generalization to OOD dataset despite lower ID Top 1 accuracy.}
\label{tab:lca_toy}
\end{table}


\section{Metric}
In this section, we outline the metrics adopted for our experiment.

\subsection{Correlation Measurement}
Correlation measurements quantify the degree of association between two variables. This can be further subdivided into linearity and ranking measurements.

\subsubsection{Linearity Measurement}
Linearity measurement evaluates the strength and direction of a linear relationship between two continuous variables. We use the R² and Pearson correlation coefficients to assess linearity.

\textbf{R² (Coefficient of determination)}:
The R², or coefficient of determination, quantifies the proportion of the variance in the dependent variable that can be predicted from the independent variable(s). It ranges from 0 to 1, where 1 indicates perfect predictability. It is defined as:
\begin{equation}
R^2 = 1 - \frac{\sum_{i=1}^{n} (y_i - f(x_i))^2}{\sum_{i=1}^{n} (y_i - \bar{y})^2}
\end{equation}
where $f(x_i)$ is the prediction of $y_i$ from the model, $\bar{y}$ is the mean of the actual $y$ values, and $n$ is the number of data points.

\textbf{PEA (Pearson correlation coefficient)}: The Pearson correlation coefficient, denoted as $r$, measures the linear relationship between two datasets. It is defined as:
\begin{equation}
r = \frac{\sum_{i=1}^{n} (x_i - \bar{x})(y_i - \bar{y})}{\sqrt{\sum_{i=1}^{n} (x_i - \bar{x})^2}\sqrt{\sum_{i=1}^{n} (y_i - \bar{y})^2}}
\end{equation}
where $\bar{x}$ and $\bar{y}$ are the mean values of the datasets $x$ and $y$, respectively, and $n$ is the number of data points.

\subsubsection{Ranking measurement}
Ranking measurement evaluates the degree of correspondence between the rankings of two variables, even when their relationship is non-linear. The Kendall and Spearman rank correlation coefficients are metrics used for this purpose.

\textbf{KEN (Kendall rank correlation coefficient)}: Also known as Kendall's tau ($\tau$), this coefficient measures the ordinal association between two variables. It is defined as:
\begin{equation}
\tau = \frac{(\text{number of concordant pairs}) - (\text{number of discordant pairs})}{\frac{1}{2} n (n - 1)}
\end{equation}
where $n$ is the number of data points.

\textbf{SPE (Spearman rank-order correlation coefficient)}: The Spearman rank-order correlation coefficient, denoted as $\rho$, assesses the monotonic relationship between two variables. It is defined as:
\begin{equation}
\rho = 1 - \frac{6\sum_{i=1}^{n} d_i^2}{n(n^2 - 1)}
\end{equation}
where $d_i$ is the difference between the ranks of corresponding data points in the two datasets and $n$ is the number of data points.

\subsection{Taxonomy Measurement}

Taxonomy measurement is designed to assess the alignment between the model-predicted class ranking and the predefined class taxonomy hierarchy tree. This is also referred to as `mistake severity' or `taxonomic distance'.  

\subsubsection{LCA distance}
Following~\citep{bertinetto2020making, valmadre2022hierarchical}, we define LCA distance using a predefined hierarchy tree, as indicated in Fig.~\ref{fig:hier}. We adopt class distance in a hierarchical tree format to denote inter-class relationships, which is necessary to calculate LCA and ELCA (cf. Section~\ref{ELCA_sup}). Given a ground-truth node y (node 1 in the plot), a model prediction node $y'$, and their lowest common ancestor class node $N_{LCA}(y,y')$. We define it as:
\begin{equation}
D_{LCA}(y', y) := f(y)-f(N_{LCA}(y,y'))
\end{equation}
where $f(\cdot)$ represents a function for a node's score, such as the tree depth or information content.

\textbf{Scores as tree depths}:
We define a function $P(x)$ to retrieve the depth of node x from tree T. Then, LCA distance is defined as:
\begin{equation}
D_{LCA}^{P}(y', y) := (P(y)-P(N_{LCA}(y',y)))+(P(y')-P(N_{LCA}(y',y))),
\end{equation}
where we also append $(P(y')-P(N_{LCA}(y',y)))$ to counter tree imbalance.

\textbf{Scores as information}:
Defining score as tree depth may be vulnerable to an imbalanced hierarchical tree. Thus, we also define a node's score as information to put more weight on nodes with more descendants. Formally, following~\citep{valmadre2022hierarchical}, we apply a uniform distribution p to all leaf nodes in the tree that indicate a class in the classification task. The probability of each intermediate node in the tree is calculated by recursively summing the scores of its descendants. Then, the information of each node is calculated as $I(node) := -log2(p)$. The LCA distance is then defined as:

\begin{equation}
D_{LCA}^{I}(y', y) := I(y)-I(N_{LCA}(y',y)),
\end{equation}

In this work, we adopt $D_{LCA}^{I}(y', y)$ for LCA measurements, and $D_{LCA}^{P}(y', y)$ for linear probing experiments.

\subsection{ELCA distance}
\label{ELCA_sup}
For a sample $X_i$ whose ground-truth class is $y_i$, and the model outputs $(\widehat{p}_{1,i},\dots,\widehat{p}_{K,i})$ over the $K$ classes (e.g., 1000 in ImageNet), we define the \textbf{Expected Lowest Common Ancestor Distance} (ELCA): $$ D_{ELCA}(model, \mathcal{M}) := \frac{1}{nK}\sum_{i=1}^n\sum_{k=1}^K \widehat{p}_{k,i} \cdot D_{LCA}(k, y_i)$$
From a probabilistic perspective, $D_{ELCA}$ is a weighted measure of mistake severity according to the model's confidence in each node in the hierarchy. Intuitively, it combines the LCA distance with a cross-entropy measurement.

The proposed ELCA distance provides a more generalized metric for assessing model performance compared to Top 1 accuracy, LCA distance, and cross entropy. Top 1 accuracy only considers the top-ranked class; LCA distance measures the Top n class rankings but treats each class equally~\citep{bertinetto2020making}; Cross-entropy solely focuses on the model's assigned probability to the ground-truth class, and ELCA extends it to all classes.
The ELCA distance captures the probabilistic distribution of mistake severity across all candidate classes.

For implementation, ELCA is a weighted combination of the LCA distance for each leaf node [1,2,3,4] as in Fig.~\ref{fig:hier}, weighted by class probability. Formally, for each prediction node $X_i$, the probabilistic distribution over all candidate classes can be obtained by applying a softmax function $softmax(x) : \mathbb{R} \rightarrow [0,1]$ to get model outputs probability $(\widehat{p}_{1,i},\dots,\widehat{p}_{K,i})$ over the $K$ classes (e.g., 1000 in ImageNet).

In Table~\ref{tab:OOD_LCA_sup}, we also demonstrate that models with better OOD generalization (OOD Top 1
accuracy) usually also have lower LCA/ELCA distances.

\begin{table*}[ht]
\small
\centering
\resizebox{\textwidth}{!}{
\begin{tabular}{ccccccccccccccccccc}
\hline
\textbf{Model} &
  \multicolumn{3}{c}{\textbf{ImageNet}} &
  \multicolumn{3}{c}{\textbf{ImageNetv2}} &
  \multicolumn{3}{c}{\textbf{ImageNet-S}} &
  \multicolumn{3}{c}{\textbf{ImageNet-R}} &
  \multicolumn{3}{c}{\textbf{ImageNet-A}} &
  \multicolumn{3}{c}{\textbf{ObjectNet}} \\ \cline{1-19} 
 &
  LCA &
  ELCA &
  \multicolumn{1}{c|}{Top1} &
  LCA &
  ELCA &
  \multicolumn{1}{c|}{Top1} &
  LCA &
  ELCA &
  \multicolumn{1}{c|}{Top1} &
  LCA &
  ELCA &
  \multicolumn{1}{c|}{Top1} &
  LCA &
  ELCA &
  \multicolumn{1}{c|}{Top1} &
  LCA &
  ELCA &
  Top1 \\
ResNet18~\cite{he2016deep} &
  6.643 &
  7.505 &
  \multicolumn{1}{c|}{0.698} &
  6.918 &
  7.912 &
  \multicolumn{1}{c|}{0.573} &
  8.005 &
  9.283 &
  \multicolumn{1}{c|}{0.202} &
  8.775 &
  8.853 &
  \multicolumn{1}{c|}{0.330} &
  8.449 &
  9.622 &
  \multicolumn{1}{c|}{0.011} &
  8.062 &
  8.636 &
  0.272 \\
ResNet50~\cite{he2016deep} &
  6.539 &
  \textbf{7.012} &
  \multicolumn{1}{c|}{\textbf{0.733}} &
  6.863 &
  \textbf{7.532} &
  \multicolumn{1}{c|}{\textbf{0.610}} &
  7.902 &
  \textbf{9.147} &
  \multicolumn{1}{c|}{0.235} &
  8.779 &
  \textbf{8.668} &
  \multicolumn{1}{c|}{0.361} &
  8.424 &
  \textbf{9.589} &
  \multicolumn{1}{c|}{0.018} &
  8.029 &
  \textbf{8.402} &
  0.316 \\ \cline{3-3} \cline{6-6} \cline{9-9} \cline{12-12} \cline{15-15} \cline{18-18}
CLIP\_RN50~\cite{radford2021learning} &
  6.327 &
  \textbf{9.375} &
  \multicolumn{1}{c|}{0.579} &
  6.538 &
  \textbf{9.442} &
  \multicolumn{1}{c|}{0.511} &
  6.775 &
  9.541 &
  \multicolumn{1}{c|}{0.332} &
  7.764 &
  9.127 &
  \multicolumn{1}{c|}{0.562} &
  7.861 &
  9.526 &
  \multicolumn{1}{c|}{0.218} &
  7.822 &
  8.655 &
  0.398 \\
CLIP\_RN50x4~\cite{radford2021learning} &
  \textbf{6.166} &
  9.473 &
  \multicolumn{1}{c|}{0.641} &
  \textbf{6.383} &
  9.525 &
  \multicolumn{1}{c|}{0.573} &
  \textbf{6.407} &
  \textbf{9.518} &
  \multicolumn{1}{c|}{\textbf{0.415}} &
  \textbf{7.435} &
  \textbf{8.982} &
  \multicolumn{1}{c|}{\textbf{0.681}} &
  \textbf{7.496} &
  \textbf{9.388} &
  \multicolumn{1}{c|}{\textbf{0.384}} &
  \textbf{7.729} &
  \textbf{8.354} &
  \textbf{0.504} \\ \hline
\end{tabular}%
}
\caption{\small
\textbf{Model performance corresponds to mistake severity.} 
\contour{red}{\textcolor{red}{LCA $\downarrow$}} / \contour{red}{\textcolor{red}{ELCA $\downarrow$}} / \contour{blue}{\textcolor{blue}{Top1 $\uparrow$}} indicate measurements on a given dataset. We present two pairs of model comparisons from the VMs and VLMs families with different generalization abilities. \textit{Note that ELCA should not be compared across modalities, as it is sensitive to logit temperature.}}
\label{tab:OOD_LCA_sup}
\end{table*}

\section{Experiment Setup}

\subsection{K-mean Clustering for Latent Class Hierarchy Construction}
\label{section_kmean_apx}
As depicted in Fig~\ref{fig:kmean}, we begin with a pretrained model $M$, in-distribution image data $X$, and labels $Y$ for $k$ classes. Initially, we extract the in-distribution data features $M(X)$. With known labels, we categorize $M(X)$ by $Y$, resulting in $k$ average class features, denoted as $k$$X$. Utilizing these per-class average features, we perform a 9-layer hierarchical clustering. For $k$$X$, we apply the K-means algorithm, setting the number of cluster centers as $2^{i}$, where $i$ ranges from ${1,2,3,4,...,9}$ since $2^9 < 1000$ (ImageNet have 1000 classes). This procedure results in 9 cluster outcomes. Subsequently, we find the LCA node between each pair of the $k$ classes, to determine the cluster level at which both classes exists in the same cluster. We use the height of the common cluster as their pairwise LCA height to be retrieved at training/evaluation. By definition, all classes share a base cluster level of 10.

\subsection{Soft Loss for Hierarchy Alignment}
\label{linear_setup}
\begin{figure*}[ht]
\centering
\resizebox{\textwidth}{!}{
\includegraphics[width=16cm]{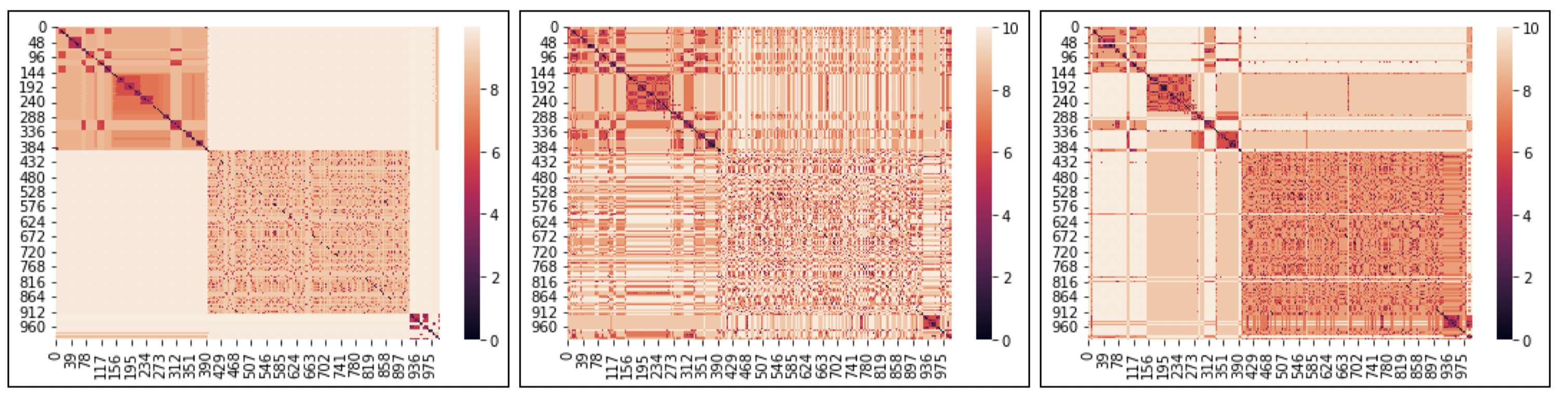}
}
\vspace{-4mm}
\caption{\textbf{Visualization of pair-wise LCA distance for ImageNet classes}. Each row signifies the LCA distance between a specific class and the reference class, arranged in ascending order, with the diagonal index indicating the shortest distance. From left to right: WordNet hierarchy; matrix constructed from ResNet50~\citep{he2016deep}; and matrix constructed from CLIP ResNet50~\citep{radford2021learning}.}
\label{fig:LCA_matrix}
\end{figure*}

This section illustrates the loss function used in our linear probing experiment. For a dataset with $n$ classes, we first establish an $n \times n$ LCA distance matrix $M$ (visualize in Figure~\ref{fig:LCA_matrix}), where $M[i, k]$ indicates the pairwise LCA distance $D_{\text{LCA}}(i, k)$, calculated using either the WordNet hierarchy or latent hierarchy derived from the K-means clustering (as introduced in the main paper). Next, we scale $M$ by applying a temperature term $T$, and finally apply MinMax scaling to normalize the values between 0 and 1. 
$$ M_{\text{LCA}} = \text{MinMax}(M^T) $$

As shown in the code snippet below, we construct the auxiliary loss by assigning class likelihoods beyond the top-1 (one-hot), extending to all classes. Similar to adopting one-hot encoding to let the model focus on the top-1 ground-truth, we use the reverse of LCA matrix as an alignment indicator, where ground-truth index have the largest value of 1. This alignment can be applied to both BCE and CE types of loss. Details in our \href{https://github.com/ElvishElvis/LCA-on-the-line}{code}. 

\begin{algorithm}
\caption{LCA Alignment Loss}
\begin{algorithmic}[1]

\Function{LCA\_Alignment\_Loss}{logits, targets, alignment\_mode, LCA\_matrix, lambda\_weight=0.03}
    \State \texttt{reverse\_LCA\_matrix} $\gets 1 - $ \texttt{LCA\_matrix}
    \State Compute predicted probabilities: \texttt{probs} $\gets$ \texttt{softmax(logits, dim=1)}
    \State One-hot encode the targets: \texttt{one\_hot\_targets}
    \State Compute standard cross-entropy loss: 
    \Statex \hspace{2em} \texttt{standard\_loss} $\gets -\sum($\texttt{one\_hot\_targets} $\cdot \log($\texttt{probs}$), \texttt{dim}=1)$

    \If{\texttt{alignment\_mode} $==$ \texttt{`BCE'}}
        \State \texttt{criterion} $\gets$ \texttt{BCEWithLogitsLoss(reduction=`none')}
        \State Compute soft loss: 
        \Statex \hspace{2em} \texttt{soft\_loss} $\gets$ \texttt{mean(criterion(logits, reverse\_LCA\_matrix[targets]), dim=1)}
    \ElsIf{\texttt{alignment\_mode} $==$ \texttt{`CE'}}
        \State Compute soft loss:
        \Statex \hspace{2em} \texttt{soft\_loss} $\gets -$ \texttt{mean(reverse\_LCA\_matrix[targets]} $\cdot \log($\texttt{probs}$), \texttt{dim}=1)$
    \EndIf
    \State \texttt{total\_loss} $\gets$ \texttt{lambda\_weight} $\cdot$ \texttt{standard\_loss} $ + $ \texttt{soft\_loss}

    \State Return mean loss over the batch: \texttt{return mean(total\_loss)}
\EndFunction
\end{algorithmic}
\end{algorithm}

For the experiments in the main paper, we set lambda=0.03, temperature=25, and use CE as the soft loss. Note that a smaller lambda scales down the standard cross-entropy loss. We found that using a large temperature, which assign semantic-closer classes with a larger likelihood, boost model generalization better.


\subsection{Ablation study: Using class ontology as soft labels}
\label{ablation}
In Table~\ref{tab:ablation}, we present ablation study on soft loss labels for linear probing from section~\ref{soft_labels_lca}.

\begin{table}[]
\centering
\resizebox{\textwidth}{!}{%
\begin{tabular}{llcccccc}
\hline
                                  &                                 & ImgNet        & ImgNet-V2     & ImgNet-S      & ImgNet-R      & ImgNet-A      & ObjectNet     \\ \hline
\multirow{6}{*}{ResNet 18~\cite{he2016deep}}        & CE-only                         & 69.4          & 56.4          & 19.7          & 31.9          & 1.1           & 27.0          \\
                                  & CE + interpolation              & 69.4          & 56.6          & 19.9          & 32.7          & 1.3           & 27.4          \\
                                  & \textbf{(Ours)} CE + Soft Loss (no ID accuracy drop)          & \textbf{69.5} & 56.5          & 19.7          & 32.4          & 1.1           & 27.3          \\
                                  & \textbf{(Ours)} CE + Soft Loss (pro-OOD)             & 69.2          & 56.4          & 20.3          & 34.1          & \textbf{1.4}  & 27.6          \\
                                  & \textbf{(Ours)} CE + Soft Loss + interpolation (no ID accuracy drop) & 69.4          & \textbf{56.9} & 20.7          & 33.8          & 1.2           & 28.0          \\
                                  & \textbf{(Ours)} CE + Soft Loss + interpolation (pro-OOD)    & 68.0          & 55.9          & \textbf{21.2} & \textbf{35.1} & \textbf{1.4}  & \textbf{28.6} \\ \hline
\multirow{6}{*}{ResNet 50~\cite{he2016deep}}        & CE-only                         & 79.5          & 67.9          & 25.5          & 36.5          & 10.3          & 43.2          \\
                                  & CE + interpolation              & 79.5          & 67.8          & 25.6          & 36.6          & 10.6          & 43.3          \\
 & \textbf{(Ours)} CE + Soft Loss (no ID accuracy drop)          & \textbf{79.8} & \textbf{68.6} & \textbf{27.7} & \textbf{42.5} & \textbf{16.2} & \textbf{45.5} \\
                                  & \textbf{(Ours)} CE + Soft Loss (pro-OOD)             & \textbf{79.8} & \textbf{68.6} & \textbf{27.7} & \textbf{42.5} & \textbf{16.2} & \textbf{45.5} \\
 & \textbf{(Ours)} CE + Soft Loss + interpolation (no ID accuracy drop) & \textbf{79.8} & \textbf{68.6} & \textbf{27.7} & \textbf{42.5} & \textbf{16.2} & \textbf{45.5} \\
 & \textbf{(Ours)} CE + Soft Loss + interpolation (pro-OOD)    & \textbf{79.8} & \textbf{68.6} & \textbf{27.7} & \textbf{42.5} & \textbf{16.2} & \textbf{45.5} \\ \hline
\multirow{6}{*}{VIT-B~\cite{dosovitskiy2020image}}            & CE-only                         & 75.8          & \textbf{62.9}          & 27.0          & 40.5          & 8.0           & 27.6          \\
                                  & CE + interpolation              & 75.7          & 62.4          & 27.0          & 40.5          & 8.2           & 27.7          \\
                                  & \textbf{(Ours)} CE + Soft Loss (no ID accuracy drop)          & 75.8          & 62.7          & 26.9          & 40.4          & 8.2           & 27.8          \\
                                  & \textbf{(Ours)} CE + Soft Loss (pro-OOD)             & 75.4          & 62.4 & \textbf{28.0} & \textbf{42.2} & \textbf{9.1}  & 27.9          \\
                                  & \textbf{(Ours)} CE + Soft Loss + interpolation (no ID accuracy drop) & \textbf{75.9} & 62.8          & 27.6          & 41.5          & 8.6           & \textbf{28.1} \\
                                  & \textbf{(Ours)} CE + Soft Loss + interpolation (pro-OOD)    & 75.4          & 62.4 & \textbf{28.0} & \textbf{42.2} & \textbf{9.1}  & 27.9          \\ \hline
\multirow{6}{*}{VIT-L~\cite{dosovitskiy2020image}}            & CE-only                         & \textbf{76.8} & 63.9          & 28.4          & 42.2          & 10.6          & 28.7          \\
                                  & CE + interpolation              & 76.7          & 64.0          & 28.3          & 42.1          & 10.9          & 28.9          \\
                                  & \textbf{(Ours)} CE + Soft Loss (no ID accuracy drop)          & \textbf{76.8} & \textbf{64.1} & 28.4          & 42.2          & 10.5          & 28.7          \\
                                  & \textbf{(Ours)} CE + Soft Loss (pro-OOD)             & 76.7          & 63.6          & \textbf{29.4} & \textbf{43.9} & \textbf{11.7} & \textbf{29.0} \\
                                  & \textbf{(Ours)} CE + Soft Loss + interpolation (no ID accuracy drop) & \textbf{76.8} & 63.8          & 29.2          & 43.6          & 11.5          & \textbf{29.0}          \\
                                  & \textbf{(Ours)} CE + Soft Loss + interpolation (pro-OOD)    & 76.7          & 63.6          & \textbf{29.4} & \textbf{43.9} & \textbf{11.7} & \textbf{29.0} \\ \hline
\multirow{6}{*}{ConvNext~\cite{liu2022convnet}}         & CE-only                         & 82.0          & 70.6          & 28.7          & 42.4          & 21.8          & 44.4          \\
                                  & CE + interpolation              & 82.0          & 70.8          & 28.8          & 42.3          & 22.2          & 44.7          \\
                                  & \textbf{(Ours)} CE + Soft Loss (no ID accuracy drop)          & 82.0          & 70.7          & 28.7          & 42.3          & 21.9          & 44.6          \\
                                  & \textbf{(Ours)} CE + Soft Loss (pro-OOD)             & 81.8          & \textbf{71.1} & \textbf{30.4} & \textbf{44.8} & \textbf{26.3} & \textbf{45.7} \\
                                  & \textbf{(Ours)} CE + Soft Loss + interpolation (no ID accuracy drop) & \textbf{82.1} & 71.0          & 30.0          & 44.3          & 25.2          & 45.5          \\
                                  & \textbf{(Ours)} CE + Soft Loss + interpolation (pro-OOD)    & 81.8          & \textbf{71.1} & \textbf{30.4} & \textbf{44.8} & \textbf{26.3} & \textbf{45.7} \\ \hline
\multirow{6}{*}{Swin Transformer~\cite{liu2021swin}} & CE-only                         & 83.1          & 72.0          & 30.3          & 43.5          & 29.5          & 48.3          \\
                                  & CE + interpolation              & 83.1          & 71.8          & 30.4          & 43.7          & 29.9          & 48.3          \\
                                  & \textbf{(Ours)} CE + Soft Loss (no ID accuracy drop)          & \textbf{83.2} & \textbf{72.0} & 31.0          & 44.2          & 30.9          & 49.0          \\
                                  & \textbf{(Ours)} CE + Soft Loss (pro-OOD)             & 83.0          & 71.8          & \textbf{31.6} & \textbf{45.5} & \textbf{33.3} & 49.4 \\
                                  & \textbf{(Ours)} CE + Soft Loss + interpolation (no ID accuracy drop) & \textbf{83.2} & 71.9          & 31.4          & 45.3          & 32.7          & \textbf{49.5}          \\
                                  & \textbf{(Ours)} CE + Soft Loss + interpolation (pro-OOD)    & 83.0          & 71.8          & \textbf{31.6} & \textbf{45.5}          & \textbf{33.3} & 49.4 \\ \hline
\end{tabular}%
}
\caption{\textbf{Ablation Study on Soft Loss Labels for Linear Probing from Section~\ref{soft_labels_lca}.} \textit{CE-only}: model trained with Cross-Entropy (CE) loss only, as a baseline; \textit{Soft Loss}: soft label loss generated from hierarchy; \textit{Interpolation}: linear interpolation in weight space between CE-only and the current method; \textit{No ID Accuracy Drop}: models that do not introduce an accuracy drop on ImageNet (ID) compared to the baseline (CE-only); \textit{Pro-OOD}: models with parameters that prefer the improvement of OOD generalization, even at the cost of a slight ID accuracy drop, to demonstrate the potential of our methods in enhancing generalization. Note that some models might be selected in multiple settings and appear in multiple rows. Results show that \textbf{1).} Incorporating soft labels significantly enhances OOD performance across all network architectures without sacrificing ID accuracy. \textbf{2).} Weight interpolation further boosts OOD generalization, particularly in models supervised with soft labels. \textbf{3).} Tuning the weight interpolation allows for a balance between maintaining ID accuracy and further improving OOD performance, demonstrating the method's flexibility and practicality.}
\label{tab:ablation}
\end{table}

\subsection{Does the Generalization Quality of the Pretrained Source Model Affect the Quality of Soft Labels? }
\label{label_quality_test}

\begin{figure}[ht]
\centering
\resizebox{\textwidth}{!}{
\includegraphics[width=17cm]{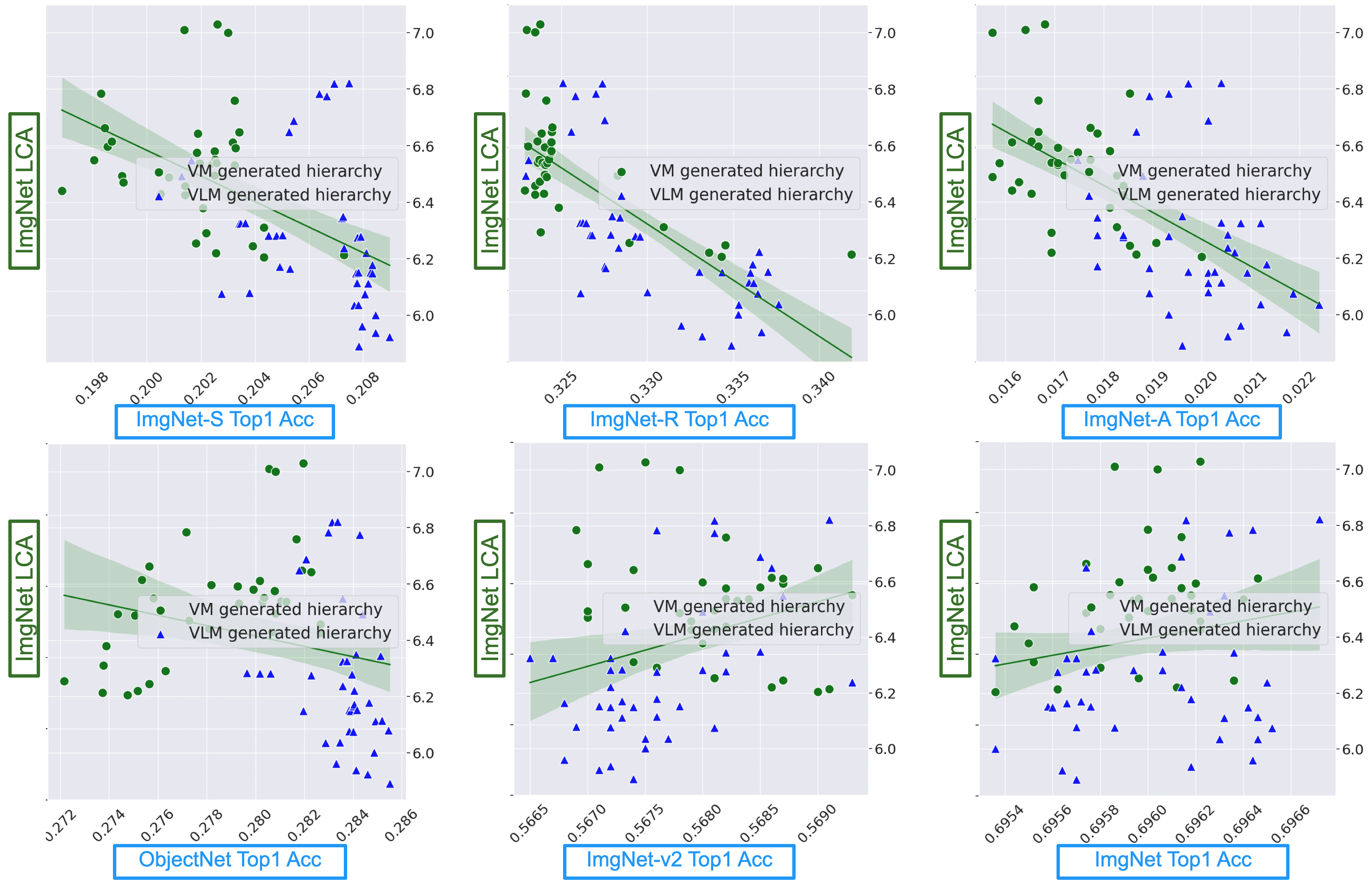}
}
\caption{\textbf{Correlation Measurement between Source Model Generalization Ability and Soft Labels Quality.} y-axis: LCA distance on ImageNet (ID dataset) between WordNet hierarchy and each of the pretrained models (that generate hierarchies).4 x-axis: top-1 accuracy on an OOD dataset by linear probing over each of the generated hierarchies. This plot visualizes the results from Table~\ref{tab:lca_distance}. It shows a moderate-strong correlation between the two variables on ImageNet-S/R/A and ObjectNet(besides some noisy data points). It also indicates that latent hierarchies constructed from VLMs tend to cluster on the right side of the x-axis, suggesting better generalization compared to those from VMs. }
\label{fig:lca_distance_matrix_all}
\end{figure}

This section continues the discussion in Section~\ref{soft_labels_lca}. We
present our findings in Table~\ref{tab:lca_distance} and Figure~\ref{fig:lca_distance_matrix_all}. The results reveal a moderate-strong correlation between 
the ID LCA of the pretrained source model, and the generalization capabilities of the linear probe model trained from the source-model-derived latent hierarchy.

\begin{table}[ht]
\small
\centering
\resizebox{\textwidth}{!}{%
\begin{tabular}{lllllll}
\hline
 &
  \textbf{ImageNet} &
  \textbf{ImageNetv2} &
  \textbf{ImageNet-S} &
  \textbf{ImageNet-R} &
  \textbf{ImageNet-A} &
  \textbf{ObjectNet} \\ \hline
\multirow{2}{*}{Corr(ID LCA, Soft Labels Quality} &
  {\ul PEA} &
  {\ul PEA} &
  {\ul PEA} &
  {\ul PEA} &
  {\ul PEA} &
  {\ul PEA} \\
 &
  0.187 &
  0.301 &
  0.535 &
  0.738 &
  0.604 &
  0.241 \\ \hline
\end{tabular}%
}
\caption{\textbf{Correlation Measurement between Source Model Generalization Ability and Soft Labels Quality.} Following the K-Means clustering algorithm, we constructed 75 LCA distance matrices (class hierarchies) from 75 pretrained source models on ImageNet. We then used these LCA distance matrices as soft labels to guide linear probing over ResNet-18 features (as described in Section~\ref{soft_labels_lca}). The table indicates a moderate-strong correlation between the in-distribution LCA of the pretrained source model and the out-of-distribution (OOD) accuracy on the linear probe model using the corresponding derived LCA distance matrix. Visualization is shown in Figure~\ref{fig:lca_distance_matrix_all}.}
\label{tab:lca_distance}
\end{table}

\subsection{Hyperparameters and Computational Resources}
In the linear probing experiment, we chose hyperparameters based on the task at hand. The learning rate was set to 0.001, with a batch size of 1024. We used the AdamW optimizer with weight decay and a cosine learning rate scheduler with a warm-up iteration. The warm-up type was set to ‘linear’ with a warm-up learning rate of 1e-5. The experiment was run for 50 epochs. For our computational resources, we utilized a single NVIDIA GeForce GTX 1080 Ti GPU.

\section{Supplementary Results}
\label{section_supplementary_result}



\subsection{Does ImageNet LCA (Taxonomic Distance) Reflect ImageNet Top-1 Accuracy?}
\label{igmNeta_to_lca_a}
\begin{figure}[ht]
\centering
\resizebox{\textwidth}{!}{
\includegraphics[width=12cm]{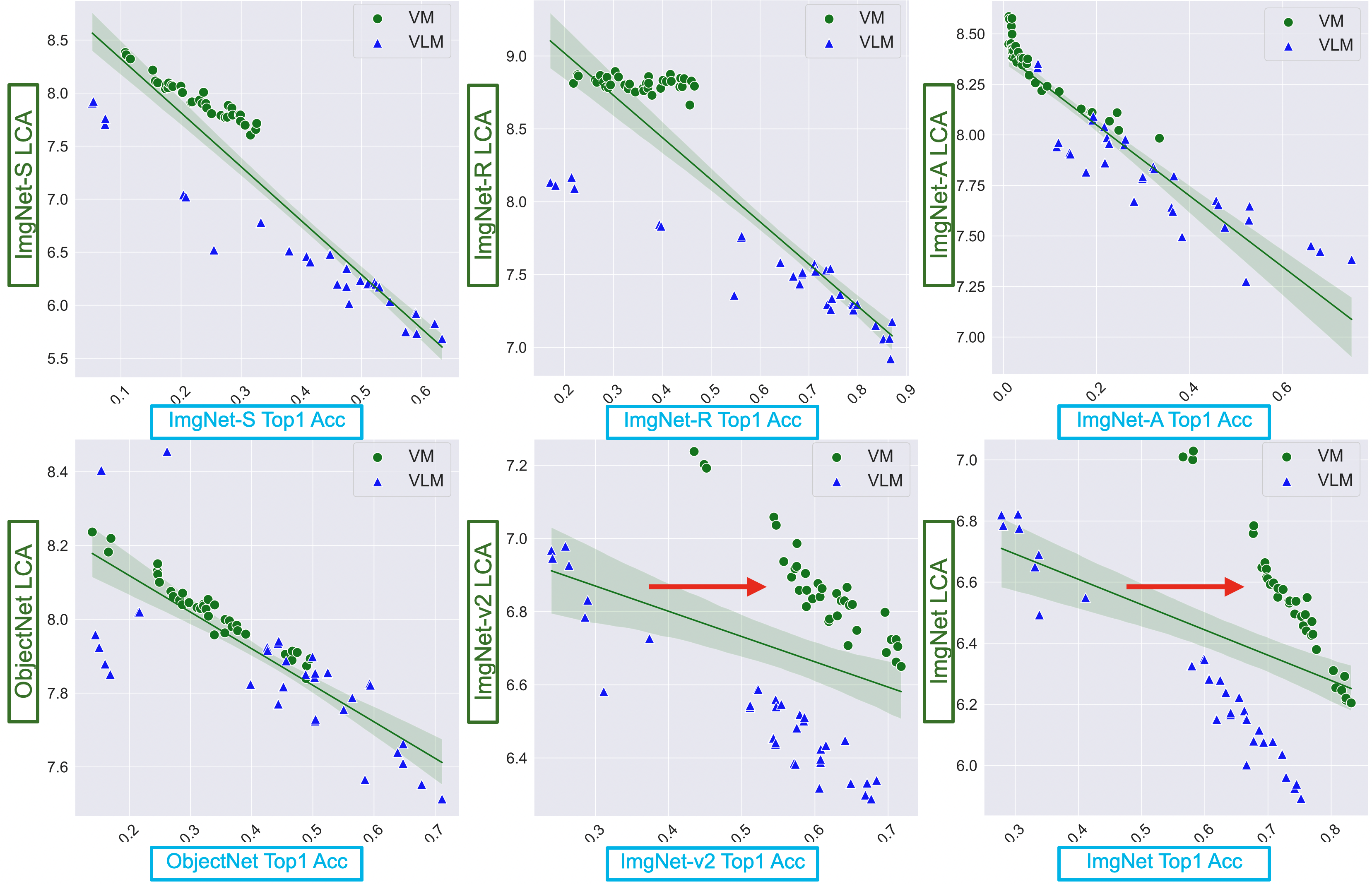}
}
\caption{\textbf{Predicting LCA (VM+VLM, 75 models) on the same dataset} As per Table~\ref{tab:correlation_ID}. Each plot's x-axis represents dataset Top-1 accuracy, while the y-axis shows LCA distance measured on the same datasets. The plots reveal that ImageNet and ImageNet-v2 do not exhibit a strong correlation between LCA and Top-1 accuracy, in contrast to other semantically distinct OOD datasets. This observation challenges the common belief that in-distribution Top-1 accuracy and LCA distance maintain the same order~\citep{deng2009imagenet,bertinetto2020making}. More details in discussion section~\ref{section_discussion}.}
\label{fig:indicator_ID}

\end{figure}

Here, we present numerical results to support the discussion in Section~\ref{section_discussion}. We challenge the common belief that LCA and Top-1 accuracy follow parallel trends within the same dataset. As illustrated in Figures~\ref{fig:indicator_ID} and Table~\ref{tab:correlation_ID}, when including both VMs and VLMs, ImageNet and ImageNet-v2 show a weak correlation between LCA and Top-1 accuracy within the same dataset. In contrast, other semantically distinct OOD datasets exhibit a stronger relationship. We provide a hypothesis in discussion section~\ref{section_discussion} on `VMs ID accuracy inflation' to explain this.

\vspace{-1mm}
\subsection{Comprehensive Results from Main Paper}
\label{sec:correlation_all_full}
Extended from Table~\ref{tab:correlation_all} and Table~\ref{tab:errorPredict} in the main paper, we present measurements on only-VMs and only-VLMs in Table~\ref{tab:correlation_all_full} and Table~\ref{tab:errorPredict_full}, respectively. Similarly, LCA is also a very good OOD indicator when involving only VMs or VLMs.

\begin{table}[ht]
\small
\centering
\resizebox{\textwidth}{!}{%
\begin{tabular}{lllllllllllll}
\hline
 &
  \multicolumn{2}{l}{\textbf{Element}} &
  \multicolumn{2}{l}{\textbf{ImageNetv2}} &
  \multicolumn{2}{l}{\textbf{ImageNet-S}} &
  \multicolumn{2}{l}{\textbf{ImageNet-R}} &
  \multicolumn{2}{l}{\textbf{ImageNet-A}} &
  \multicolumn{2}{l}{\textbf{ObjectNet}} \\
  \cline{2-13}
 &
  ID &
  OOD &
  $R^2$ &
  PEA &
  $R^2$ &
  PEA &
  $R^2$ &
  PEA &
  $R^2$ &
  PEA &
  $R^2$ &
  PEA \\ \hline
\multirow{4}{*}{ALL} &
  Top1 &
  Top1 &
  \textbf{0.962} &
  \textbf{0.980} &
  0.075 &
  0.275 &
  0.020 &
  0.140 &
  0.009 &
  0.094 &
  0.273 &
  0.522 \\
 &
  LCA &
  Top1 &
  0.339 &
  0.582 &
  \textbf{0.816} &
  \textbf{0.903} &
  \textbf{0.779} &
  \textbf{0.883} &
  \textbf{0.704} &
  \textbf{0.839} &
  \textbf{0.915} &
  \textbf{0.956} \\ \cline{2-13} 
 &
  Top1 &
  Top5 &
  \textbf{0.889} &
  \textbf{0.943} &
  0.052 &
  0.229 &
  0.004 &
  0.060 &
  0.013 &
  0.115 &
  0.262 &
  0.512 \\
 &
  LCA &
  Top5 &
  0.445 &
  0.667 &
  \textbf{0.811} &
  \textbf{0.901} &
  \textbf{0.738} &
  \textbf{0.859} &
  \textbf{0.799} &
  \textbf{0.894} &
  \textbf{0.924} &
  \textbf{0.961} \\ \hline
\multirow{4}{*}{VLM} &
  Top1 &
  Top1 &
  \textbf{0.996} &
  \textbf{0.998} &
  \textbf{0.860} &
  \textbf{0.927} &
  0.851 &
  0.923 &
  0.578 &
  0.761 &
  \textbf{0.945} &
  \textbf{0.972} \\
 &
  LCA &
  Top1 &
  0.956 &
  0.978 &
  0.850 &
  0.921 &
  \textbf{0.867} &
  \textbf{0.931} &
  \textbf{0.691} &
  \textbf{0.832} &
  0.936 &
  0.968 \\ \cline{2-13} 
 &
  Top1 &
  Top5 &
  \textbf{0.988} &
  \textbf{0.994} &
  \textbf{0.867} &
  \textbf{0.931} &
  0.820 &
  0.906 &
  0.740 &
  0.860 &
  \textbf{0.970} &
  \textbf{0.985} \\
 &
  LCA &
  Top5 &
  0.930 &
  0.964 &
  0.852 &
  0.923 &
  \textbf{0.826} &
  \textbf{0.909} &
  \textbf{0.822} &
  \textbf{0.906} &
  0.931 &
  0.965 \\ \hline
\multirow{4}{*}{VM} &
  Top1 &
  Top1 &
  \textbf{0.996} &
  \textbf{0.998} &
  \textbf{0.824} &
  \textbf{0.908} &
  \textbf{0.801} &
  \textbf{0.895} &
  0.523 &
  0.723 &
  0.900 &
  0.949 \\
 &
  LCA &
  Top1 &
  0.976 &
  0.988 &
  0.798 &
  0.893 &
  0.768 &
  0.877 &
  \textbf{0.549} &
  \textbf{0.741} &
  \textbf{0.908} &
  \textbf{0.953} \\ \cline{2-13} 
 &
  Top1 &
  Top5 &
  \textbf{0.993} &
  \textbf{0.997} &
  \textbf{0.829} &
  \textbf{0.910} &
  \textbf{0.821} &
  \textbf{0.906} &
  0.696 &
  0.834 &
  0.919 &
  0.959 \\
 &
  LCA &
  Top5 &
  0.970 &
  0.985 &
  0.797 &
  0.893 &
  0.777 &
  0.882 &
  \textbf{0.708} &
  \textbf{0.841} &
  \textbf{0.920} &
  \textbf{0.960} \\ \hline
\end{tabular}%
}
\caption{\small
\textbf{Correlation measurement of ID LCA/Top1 with OOD Top1/Top5} on 75 models across modality following Fig~\ref{fig:indicator}.
The `ALL grouping' demonstrates that LCA has a strong correlation with OOD performance on all datasets (except ImageNet-v2).
We take the absolute value of all correlations for simplicity.
Equivalently, LCA is also a very good OOD indicator when only involved VM or VLM.}
\label{tab:correlation_all_full}
\end{table}

\begin{table}[ht]
\small
\centering
\resizebox{\textwidth}{!}{%
\begin{tabular}{lllllll}
\hline
 &
   &
  \textbf{ImageNetv2} &
  \textbf{ImageNet-S} &
  \textbf{ImageNet-R} &
  \textbf{ImageNet-A} &
  \textbf{ObjectNet} \\ \hline
ALL &
  ID Top1~\citep{miller2021accuracy} &
  \textbf{0.040} &
  0.230 &
  0.277 &
  0.192 &
  0.178 \\
    & AC~\citep{hendrycks2016baseline} & \underline{0.043}    & \underline{0.124}    & \textbf{0.113} & 0.324          & \underline{0.127} \\
 &
  Aline-D~\citep{baek2022agreement} &
  0.121 &
  0.270 &
  0.167 &
  0.409 &
  0.265 \\
 &
  Aline-S~\citep{baek2022agreement} &
  0.072 &
  0.143 &
  0.201 &
  \underline{0.165} &
  0.131 \\
 &
  (Ours) ID LCA &
  0.162 &
  \textbf{0.093} &
  \underline{0.114} &
  \textbf{0.103} &
  \textbf{0.048} \\ \hline
VLM & ID~\citep{miller2021accuracy}    & \textbf{0.014} & 0.077          & \underline{0.064}    & 0.127          & \underline{0.052} \\
    & AC~\citep{hendrycks2016baseline} & \underline{0.029}    & \textbf{0.050} & \textbf{0.044} & 0.217          & 0.088       \\
 &
  Aline-D~\citep{baek2022agreement} &
  0.151 &
  0.250 &
  0.081 &
  0.296 &
  0.260 \\
 &
  Aline-S~\citep{baek2022agreement} &
  0.070 &
  \underline{0.069} &
  0.068 &
  \textbf{0.080} &
  0.153 \\
 &
  (Ours) ID LCA &
  0.047 &
  0.083 &
  0.070 &
  \underline{0.105} &
  \textbf{0.043} \\ \hline
VM  & ID~\citep{miller2021accuracy}    & \textbf{0.013} & \textbf{0.099} & \underline{0.108}    & \textbf{0.143} & \underline{0.068} \\
 &
  AC~\citep{hendrycks2016baseline} &
  0.059 &
  0.204 &
  0.188 &
  0.441 &
  0.168 \\
 &
  Aline-D~\citep{baek2022agreement} &
  0.083 &
  0.427 &
  0.313 &
  0.665 &
  0.364 \\
 &
  Aline-S~\citep{baek2022agreement} &
  0.105 &
  0.182 &
  \textbf{0.092} &
  0.574 &
  0.216 \\
 &
  (Ours) ID LCA &
  \underline{0.029} &
  \underline{0.102} &
  0.113 &
  \underline{0.145} &
  \textbf{0.065} \\ \hline
\end{tabular}%
}
\caption{\small
\textbf{Error Prediction of OOD Datasets} across 75 models of diverse settings with \contour{red}{\textcolor{red}{MAE loss $\downarrow$}}.
Top1 in \textbf{bold} and Top2 in \underline{underline}. 
Despite ImageNet's in-distribution accuracy maintain as a significant indicator of ImageNet-v2 accuracy, the in-distribution LCA outperforms it as a robust error predictor across four naturally distributed OOD datasets.}
\label{tab:errorPredict_full}
\end{table}

\begin{table}[ht]
\small
\centering
\resizebox{\textwidth}{!}{%
\begin{tabular}{llllllllllllll}
\hline
\textbf{Model} &
  \textbf{Group} &
  \multicolumn{2}{l}{\textbf{ImageNet}} &
  \multicolumn{2}{l}{\textbf{ImageNetv2}} &
  \multicolumn{2}{l}{\textbf{ImageNet-S}} &
  \multicolumn{2}{l}{\textbf{ImageNet-R}} &
  \multicolumn{2}{l}{\textbf{ImageNet-A}} &
  \multicolumn{2}{l}{\textbf{ObjectNet}} \\ \hline
\multirow{12}{*}{Top1-\textgreater LCA} &
  \multirow{4}{*}{ALL} &
  {\ul $R^2$} &
  {\ul \textit{PEA}} &
  {\ul $R^2$} &
  {\ul \textit{PEA}} &
  {\ul $R^2$} &
  {\ul \textit{PEA}} &
  {\ul $R^2$} &
  {\ul \textit{PEA}} &
  {\ul $R^2$} &
  {\ul \textit{PEA}} &
  {\ul $R^2$} &
  {\ul \textit{PEA}} \\
 &
   &
  0.174 &
  0.417 &
  0.114 &
  0.337 &
  \textbf{0.835} &
  \textbf{0.914} &
  \textbf{0.770} &
  \textbf{0.878} &
  \textbf{0.851} &
  \textbf{0.923} &
  \textbf{0.657} &
  \textbf{0.810} \\
 &
   &
  {\ul \textit{KEN}} &
  {\ul \textit{SPE}} &
  {\ul \textit{KEN}} &
  {\ul \textit{SPE}} &
  {\ul \textit{KEN}} &
  {\ul \textit{SPE}} &
  {\ul \textit{KEN}} &
  {\ul \textit{SPE}} &
  {\ul \textit{KEN}} &
  {\ul \textit{SPE}} &
  {\ul \textit{KEN}} &
  {\ul \textit{SPE}} \\
 &
   &
  0.280 &
  0.266 &
  0.237 &
  0.294 &
  \textbf{0.818} &
  \textbf{0.926} &
  \textbf{0.621} &
  \textbf{0.803} &
  \textbf{0.825} &
  \textbf{0.951} &
  \textbf{0.673} &
  \textbf{0.823} \\ \cline{2-14} 
 &
  \multirow{4}{*}{VLM} &
  {\ul $R^2$} &
  {\ul PEA} &
  {\ul $R^2$} &
  {\ul PEA} &
  {\ul $R^2$} &
  {\ul PEA} &
  {\ul $R^2$} &
  {\ul PEA} &
  {\ul $R^2$} &
  {\ul PEA} &
  {\ul $R^2$} &
  {\ul PEA} \\
 &
   &
  \textbf{0.938} &
  \textbf{0.969} &
  \textbf{0.891} &
  \textbf{0.944} &
  \textbf{0.945} &
  \textbf{0.972} &
  \textbf{0.878} &
  \textbf{0.937} &
  \textbf{0.725} &
  \textbf{0.851} &
  0.510 &
  \textbf{0.714} \\
 &
   &
  {\ul KEN} &
  {\ul SPE} &
  {\ul KEN} &
  {\ul SPE} &
  {\ul KEN} &
  {\ul SPE} &
  {\ul KEN} &
  {\ul SPE} &
  {\ul KEN} &
  {\ul SPE} &
  {\ul KEN} &
  {\ul SPE} \\
 &
   &
  \textbf{0.880} &
  \textbf{0.969} &
  \textbf{0.799} &
  \textbf{0.881} &
  \textbf{0.864} &
  \textbf{0.963} &
  \textbf{0.753} &
  \textbf{0.902} &
  \textbf{0.689} &
  \textbf{0.869} &
  0.529 &
  \textbf{0.720} \\ \cline{2-14} 
 &
  \multirow{4}{*}{VM} &
  {\ul $R^2$} &
  {\ul PEA} &
  {\ul $R^2$} &
  {\ul PEA} &
  {\ul $R^2$} &
  {\ul PEA} &
  {\ul $R^2$} &
  {\ul PEA} &
  {\ul $R^2$} &
  {\ul PEA} &
  {\ul $R^2$} &
  {\ul PEA} \\
 &
   &
  \textbf{0.973} &
  \textbf{0.986} &
  \textbf{0.890} &
  \textbf{0.943} &
  \textbf{0.934} &
  \textbf{0.966} &
  0.095 &
  0.310 &
  \textbf{0.840} &
  \textbf{0.916} &
  \textbf{0.948} &
  \textbf{0.974} \\
 &
   &
  {\ul KEN} &
  {\ul SPE} &
  {\ul KEN} &
  {\ul SPE} &
  {\ul KEN} &
  {\ul SPE} &
  {\ul KEN} &
  {\ul SPE} &
  {\ul KEN} &
  {\ul SPE} &
  {\ul KEN} &
  {\ul SPE} \\
 &
   &
  \textbf{0.911} &
  \textbf{0.980} &
  \textbf{0.758} &
  \textbf{0.910} &
  \textbf{0.854} &
  \textbf{0.963} &
  0.149 &
  0.222 &
  \textbf{0.839} &
  \textbf{0.952} &
  \textbf{0.854} &
  \textbf{0.960} \\ \hline
\end{tabular}%
}
\caption{\textbf{Correlation Measurement between Top-1 Accuracy and LCA on the Same Dataset}. This analysis uses 75 models across different modalities (36 VMs and 39 VLMs) on all six ImageNet datasets. While the main paper employs ID LCA to predict OOD performance (\textit{e.g., Corr(ImageNet LCA, ImageNet-A Top-1 Accuracy)}), this setting differs by using LCA to predict Top-1 accuracy on the same dataset (\textit{e.g., Corr(ImageNet-A LCA, ImageNet-A Top-1 Accuracy)}). Following Figure~\ref{fig:indicator_ID}, we highlight strong correlation indications. For simplicity, we take the absolute value of all correlations. More details in discussion section~\ref{section_discussion}.}
\label{tab:correlation_ID}
\end{table}

{
\setlength{\tabcolsep}{0.6em} 
\begin{table*}[t]
\small
\scalebox{0.8}
{
\begin{tabular}{ccccccccccccc}
\hline
\multirow{2}{*}{Model} &
  \multicolumn{2}{c}{\textbf{ImgN}} &
  \multicolumn{2}{c}{\textbf{ImgN-v2}} &
  \multicolumn{2}{c}{\textbf{ImgN-S}} &
  \multicolumn{2}{c}{\textbf{ImgN-R}} &
  \multicolumn{2}{c}{\textbf{ImgN-A}} &
  \multicolumn{2}{c}{\textbf{ObjNet}} \\ \cline{2-13} 
               & Top1 $\uparrow$ & Test CE $\downarrow$  & Top1 $\uparrow$ & Test CE $\downarrow$ & Top1 $\uparrow$ & Test CE $\downarrow$ & Top1 $\uparrow$ & Test CE $\downarrow$ & Top1 $\uparrow$ & Test CE $\downarrow$ & Top1 $\uparrow$ & Test CE $\downarrow$ \\ \cline{2-13} 
Baseline       & 0.589           & 9.322 & 0.517           & 9.384 & 0.379           & 9.378 & 0.667           & 8.790 & 0.294           & 9.358 & 0.394           & 8.576 \\
Stack Parent   & 0.381           & 9.389 & 0.347           & 9.395 & 0.219           & 9.561 & 0.438           & 9.258 & 0.223           & 9.364 & 0.148           & 9.076 \\
Shuffle Parent & 0.483           & 9.679 & 0.432           & 9.696 & 0.329           & 9.718 & 0.557           & 9.281 & 0.236           & 9.586 & 0.329           & 8.785 \\
Taxonomy Parent &
  \textbf{0.626} &
  \textbf{9.102} &
  \textbf{0.553} &
  \textbf{9.165} &
  \textbf{0.419} &
  \textbf{9.319} &
  \textbf{0.685} &
  \textbf{8.658} &
  \textbf{0.319} &
  \textbf{9.171} &
  \textbf{0.431} &
  \textbf{8.515} \\ \hline
\end{tabular}
}
\vspace{-2mm}
\caption{\textbf{Accuracy on OOD dataset by enforcing class taxonomy:} \textbf{Baseline}: \textit{<dalmatian>}; \textbf{Stack Parent}: \textit{<dalmatian, dog, animal>}; \textbf{Taxonomy Parent}:\textit{<dalmatian, which is type of a dog, which is type of an animal>}; \textbf{Shuffle Parent}: \textit{<dalmatian, which is type of an organism, which is type of a seabird>}; The Taxonomy Parent method, which includes the full hierarchical relationship, yields the best performance, highlighting the effectiveness of incorporating structured knowledge into model predictions.}
\label{tab:parent}
\end{table*}
}

\vspace{-2mm}
\subsection{Ranking Measurement of LCA-on-the-Line}

Here we present the numerical results for ranking measures in comparison to the commonly used Top-1 in-distribution accuracy in Table~\ref{tab:correlation_all_rank}. Similarly, in-distribution LCA distance presents strong results in both preserving linearity and ranking.

\begin{table}[ht]
\small
\centering
\resizebox{\textwidth}{!}{%
\begin{tabular}{lllllllllllll}
\hline
 &
  \multicolumn{2}{l}{\textbf{Element}} &
  \multicolumn{2}{l}{\textbf{ImageNetv2}} &
  \multicolumn{2}{l}{\textbf{ImageNet-S}} &
  \multicolumn{2}{l}{\textbf{ImageNet-R}} &
  \multicolumn{2}{l}{\textbf{ImageNet-A}} &
  \multicolumn{2}{l}{\textbf{ObjectNet}} \\ \cline{2-13} 
 &
  ID &
  OOD &
  KEN &
  SPE &
  KEN &
  SPE &
  KEN &
  SPE &
  KEN &
  SPE &
  KEN &
  SPE \\ \hline
\multirow{4}{*}{ALL} &
  Top1 &
  Top1 &
  \textbf{0.840} &
  \textbf{0.947} &
  0.170 &
  0.092 &
  0.146 &
  0.042 &
  0.068 &
  0.037 &
  0.317 &
  0.339 \\
 &
  LCA &
  Top1 &
  0.421 &
  0.517 &
  \textbf{0.779} &
  \textbf{0.923} &
  \textbf{0.761} &
  \textbf{0.911} &
  \textbf{0.730} &
  \textbf{0.888} &
  \textbf{0.867} &
  \textbf{0.967} \\ \cline{2-13} 
 &
  Top1 &
  Top5 &
  \textbf{0.672} &
  \textbf{0.818} &
  0.151 &
  0.059 &
  0.134 &
  0.004 &
  0.108 &
  0.021 &
  0.279 &
  0.297 \\
 &
  LCA &
  Top5 &
  0.571 &
  0.729 &
  \textbf{0.768} &
  \textbf{0.919} &
  \textbf{0.752} &
  \textbf{0.897} &
  \textbf{0.755} &
  \textbf{0.908} &
  \textbf{0.861} &
  \textbf{0.966} \\ \hline
\multirow{4}{*}{VLM} &
  Top1 &
  Top1 &
  \textbf{0.971} &
  \textbf{0.997} &
  \textbf{0.840} &
  \textbf{0.936} &
  \textbf{0.864} &
  \textbf{0.943} &
  0.753 &
  0.915 &
  \textbf{0.905} &
  \textbf{0.982} \\
 &
  LCA &
  Top1 &
  0.882 &
  0.972 &
  0.729 &
  0.861 &
  0.762 &
  0.886 &
  \textbf{0.800} &
  \textbf{0.942} &
  0.870 &
  0.972 \\ \cline{2-13} 
 &
  Top1 &
  Top5 &
  \textbf{0.908} &
  0.980 &
  \textbf{0.848} &
  \textbf{0.951} &
  \textbf{0.882} &
  \textbf{0.959} &
  0.753 &
  0.910 &
  \textbf{0.842} &
  \textbf{0.964} \\
 &
  LCA &
  Top5 &
  0.900 &
  \textbf{0.981} &
  0.746 &
  0.879 &
  0.775 &
  0.907 &
  \textbf{0.794} &
  \textbf{0.943} &
  0.829 &
  0.955 \\ \hline
\multirow{4}{*}{VM} &
  Top1 &
  Top1 &
  \textbf{0.948} &
  \textbf{0.993} &
  \textbf{0.771} &
  \textbf{0.901} &
  \textbf{0.743} &
  \textbf{0.887} &
  \textbf{0.735} &
  \textbf{0.877} &
  \textbf{0.822} &
  \textbf{0.927} \\
 &
  LCA &
  Top1 &
  0.910 &
  0.981 &
  0.740 &
  0.882 &
  0.705 &
  0.862 &
  0.741 &
  0.851 &
  0.790 &
  0.918 \\ \cline{2-13} 
 &
  Top1 &
  Top5 &
  \textbf{0.939} &
  \textbf{0.992} &
  \textbf{0.752} &
  \textbf{0.894} &
  \textbf{0.758} &
  \textbf{0.901} &
  \textbf{0.818} &
  \textbf{0.941} &
  \textbf{0.815} &
  \textbf{0.920} \\
 &
  LCA &
  Top5 &
  0.894 &
  0.977 &
  0.733 &
  0.879 &
  0.707 &
  0.871 &
  0.780 &
  0.916 &
  0.783 &
  0.911 \\ \hline
\end{tabular}%
}
\caption{\textbf{Ranking measurement of ID LCA/Top1 with OOD Top1/Top5} on 75 models across modality(36 VMs and 39 VLMs); As shown in the `ALL grouping', LCA shows a much better result in preserving the model relative ranking to model OOD performance on all OOD datasets (with the exception of ImageNet-v2), which indicates its superiority for model selection.}
\label{tab:correlation_all_rank}
\end{table}



\end{document}